\documentclass{article}

\PassOptionsToPackage{numbers, compress}{natbib}

\usepackage[preprint]{neurips_2020}

\usepackage[utf8]{inputenc} 
\usepackage[T1]{fontenc}    
\usepackage{hyperref}       
\usepackage{url}            
\usepackage{booktabs}       
\usepackage{amsfonts}       
\usepackage{nicefrac}       
\usepackage{microtype}      

\usepackage{graphicx}
\usepackage{enumitem}
\usepackage{wrapfig}
\usepackage{lipsum}
\usepackage{booktabs}

\newcommand{\ie}{\textit{i.e.}}
\newcommand{\eg}{\textit{e.g.}}

\title{Adversarial Visual Robustness by Causal Intervention}

\author{
\textbf{Kaihua Tang}\\
Nanyang Technological University\\
{\tt\small kaihua001@e.ntu.edu.sg}

\and
\textbf{Mingyuan Tao}\\
Alibaba Group\\
{\tt\small juchen.tmy@alibaba-inc.com}

\and
\textbf{Hanwang Zhang}\\
Nanyang Technological University\\
{\tt\small hanwangzhang@ntu.edu.sg}
}

\begin{document}

\maketitle

\begin{abstract}
Adversarial training is the de facto most promising defense against adversarial examples. Yet, its passive nature inevitably prevents it from being immune to unknown attackers. To achieve a proactive defense, we need a more fundamental understanding of adversarial examples, beyond the popular bounded threat model. In this paper, we provide a causal viewpoint of adversarial vulnerability: the cause is the spurious correlation ubiquitously existing in learning, \ie, the confounding effect, where attackers are precisely exploiting these effects. Therefore, a fundamental solution for adversarial robustness is by causal intervention. As these visual confounders are imperceptible in general, we propose to use the instrumental variable that achieves causal intervention without the need for confounder observation. We term our robust training method as Causal intervention by instrumental Variable (CiiV). It's a causal regularization that 1) augments the image with multiple retinotopic centers and 2) encourages the model to learn causal features, rather than local confounding patterns, by favoring features linearly responding to spatial interpolations. Extensive experiments on a wide spectrum of attackers and settings applied in CIFAR-10, CIFAR-100, and mini-ImageNet demonstrate that CiiV is robust to adaptive attacks, including the recent AutoAttack. Besides, as a general causal regularization, it can be easily plugged into other methods to further boost the robustness.\footnote{Codes are available at the following Github project:  
\url{https://github.com/KaihuaTang/CiiV-Adversarial-Robustness.pytorch}}
\end{abstract}

\section{Introduction}
\label{sec.1}

Despite the remarkable progress achieved by Deep Neural Networks (DNNs), adversarial vulnerability~\cite{gdfl2014explaining} keeps haunting the computer vision community since it has been spotted by \cite{szegedy2013intriguing}. Over the years, we have witnessed many defenders, who claim to be ``well-rounded'', were soon found to lack fair benchmarking, \eg, adaptive adversary~\cite{croce2020reliable, tramer2020adaptive}, or misconduct the attack, \eg, obfuscated gradient~\cite{athalye2018obfuscated}. Therefore, the most promising defender remains to be the intuitive Adversarial Training and its variants~\cite{kannan2018adversarial, cui2020learnable}. Due to the ``training'' nature, its adversarial robustness is largely dependent on the knowledge of attackers and whether the training set contains sufficient adversarial samples from various attackers as many as possible~\cite{tramer2017ensemble}, yet, brute-forcely enumerating all attackers is prohibitively expensive, making adversarial training mainly over-fitting to known attackers~\cite{schott2018towards}. What's worse, in few-/zero-shot scenarios, it is even impossible to collect enough adversarial training samples based on the out-of-distribution/unseen samples~\cite{zhang2019limitations}.

In other words, adversarial training is a ``passive immunization'', which cannot react to the ever-evolving attacks responsively. To proactively achieve adversarial robustness, we have to find the ``origin'' of adversarial perturbations. Previous methods blame adversarial vulnerability on the inherent flaws in fitting models to the limited high-dimensional data~\cite{gdfl2014explaining, gilmer2018adversarial,schmidt2018adversarially}. However, simply regarding adversarial samples as ``bugs'' cannot explain their well-generalizing behaviors~\cite{xie2020adversarial, ilyas2019adversarial}. Recent studies~\cite{salman2020adversarially, ilyas2019adversarial} show that adversarial examples are not ``bugs'' but \emph{predictive} features that can only be exploited by machines. Such results urge us to investigate the essential difference between machines and humans.

However, we believe that it is too early for us to shirk responsibility and leave it to the ever-elusive open problem before we answer the following two key questions:

\noindent\textbf{Q1: \textit{What are the non-robust but predictive features?}} \cite{ilyas2019adversarial} use adversarial examples to distinguish the robust and non-robust features. However, this will only allow us to recognize the non-robust features as ``adversarial perturbations'' again, which is, unfortunately, circular reasoning. Therefore, we need a fundamental yet different angle to define the robustness of features beyond the conventional adversarial one.

\noindent\textbf{Q2: \textit{Why do complex systems (human vision) ignore these predictive features that simple systems (DNNs) can capture instead?}} Given the fact that biological visions are more complex than machines in terms of both neuron amount~\cite{herculano2012remarkable} and diversity~\cite{masland2001fundamental, kim2020modeling}, there is no reason for human vision to extract ``less feature'' than machines. Therefore, there must be a mechanism in human vision that deliberately ignores these features.

\begin{figure}[t]
   \begin{center}
   \includegraphics[width=140mm]{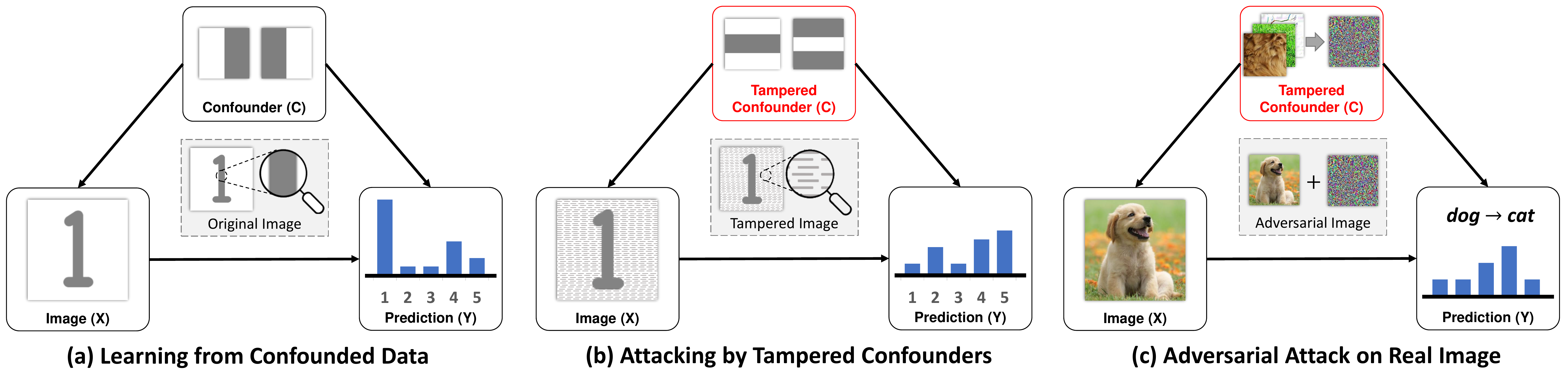}
   \end{center}
   \vspace{-2mm}
   \caption{(a) A digit classifier confounded by counting edges. (b) Attacking the model through tampered confounders. (c) Constructing adversarial perturbations through an ensemble of tampered confounders, \eg, local textures, small edges, and faint shadows.}
   \label{fig:1.1}
   \vspace{-5mm}
\end{figure}

In this paper, we answer the above two questions from a causal perspective~\cite{pearl2009causality}---a powerful lens seeing through the generative nature of adversarial attacks. For \textbf{Q1}, we postulate that non-robust features are confounding effects, which are spurious correlations established by related but non-causal features. Take Figure~\ref{fig:1.1} (a) as an intuitive example, where a large number of vertical edges co-occur with the digit ``1''. As a result, a model trained by associating samples with labels will recklessly use the counting of vertical edges---the confounding effect---as the indicator of digit ``1'' without learning the overall causal structure. Therefore, once tampered edges are constructed, which is much easier than editing the entire digit directly, the confounding effect will mislead the model prediction as shown in Figure~\ref{fig:1.1} (b).

In general, any pattern co-occurred with certain labels can constitute confounders. Most of them are even imperceptible, like local textures, small edges, and faint shadows. Since DNN models are based on the statistical association between input and output, they inevitably learn these spurious correlations, which are ``predictive'' when the distribution of confounders remains the same in training and testing. However, their brittle nature makes them vulnerable to small perturbations as shown in Figure~\ref{fig:1.1} (c). In Section~\ref{sec.4}, we will provide a formal revisit for the adversarial attack in the causal viewpoint, where we also design a Confounded-Toy dataset to demonstrate how an adversarial attacker fools the model by exploiting the confounding effect.

\begin{figure}[t]
   \begin{minipage}[b]{1.0\linewidth}
   \centerline{\includegraphics[width=100mm]{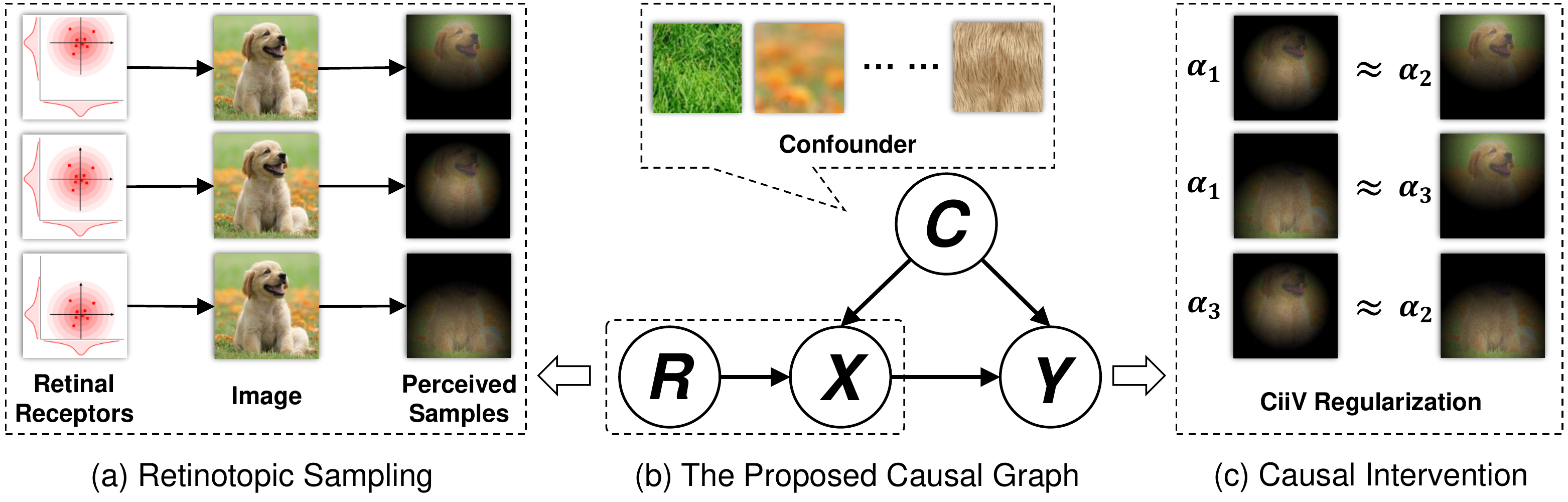}}
   \end{minipage}
   \caption{The proposed CiiV framework (detailed in Section~\ref{sec.6}): (a) the retinotopic augmentation that serves as the instrumental variable; (b) the proposed causal graph; (c) the causal intervention made by the proposed regularization that suppresses non-robust confounding effects.}
   \label{fig:1.2}
   \vspace{-5mm}
\end{figure}

Unlike machine vision that scans all the pixels in an image at once, human vision continuously perceives the image using ``retinotopic sampling''~\cite{arcaro2009retinotopic} via non-uniformly distributed retinal photoreceptors at each time frame as shown in Figure~\ref{fig:1.2} (a). We conjecture that such a mechanism is the answer to \textbf{Q2}, because it can be viewed as causal intervention by using instrumental variable~\cite{greenland2000introduction}, denoted as $R$ in Figure~\ref{fig:1.2} (b). With the help of $R$, the confounded image observation $X$ is no longer dictated only by the confounders. Since the choice of $R$ is designed to be independent of $C$, as it only depends on the structure of retina, its direct effect on $Y$ can thus be used to mitigate the confounding effect even though $C$ is unobserved. Intuitively, non-robust confounder patterns are local impulses that won't perform consistently across different retinotopic centers. They are either captured or not by a retinotopic observation. Meanwhile, causal features are consistent structures. Forcing a model to learn features that linearly vary with the change of $R$ can suppress unstable confounding effects. To this end, in Section~\ref{sec.6}, we propose the Causal intervention by instrumental Variable (CiiV $\backslash$si:v$\backslash$) framework that combines a spatial data augmentation through retinotopic sampling with a consistency regularization loss as shown in Figure~\ref{fig:1.2} (c).

Our key contributions are as follows: 
\begin{itemize}[leftmargin=1.2em]
  \item We introduce a causal regularization termed CiiV to suppress the learning of non-robust features in DNN models, which not only offers a proactive defender, but also opens a novel yet fundamental viewpoint of adversary research.

  \item Extensive experiments on a wide range of settings from the adversarial evaluation checklist~\cite{carlini2019evaluating} in CIFAR-10, CIFAR-100, and mini-ImageNet demonstrate that CiiV can withstand adaptive attacks, including the state-of-the-art AutoAttack~\cite{croce2020reliable}. 
  
  \item As a general regularization that is orthogonal to most of the previous defenders, the proposed CiiV can be easily plugged into other methods to further boost their adversarial robustness.
\end{itemize}

\section{Related Work}
\label{sec.2}

\noindent\textbf{Adversarial Examples.} 
Adversarial examples undermine the reliability and interpretability of DNN models in various domains~\cite{wang2021understanding, qi2021stabilized, finlayson2019adversarial, xie2017adversarial, xiang2019generating, cisse2017houdini, carlini2018audio, huang2017adversarial} and settings~\cite{diao2021basarblackbox, moosavi2016deepfool, kurakin2016adversarial, papernot2016limitations, carlini2017towards, zheng2019dist}. Despite of various defenders proposed to improve the adversarial robustness, a universal remedy that can proactively defend against all the known and unknown attackers is still absent. Generally, the existing defenders fall into the following four categories: adversarial training~\cite{szegedy2013intriguing, wong2019fast, dong2020api}, data augmentation~\cite{zhang2018mixup}, de-noising~\cite{buckman2018thermometer, xie2019feature}, and certified defense~\cite{cohen2019certified}. In Section~\ref{sec.5} we will systematically revisit them and compare them to the proposed CiiV from a causal viewpoint.

\noindent\textbf{Causality in Adversarial Robustness.}
Recently, causality has gradually been accepted as a potential way to explain adversarial robustness. \cite{zhang2020causal, zhang2021adversarial} provide a causal perspective to understand the adversarial vulnerability of DNN models; \cite{yang2019causal} utilize the supervised pixel-wise masking to conduct causal intervention; \cite{singh2021learning} attempt to unify the adversarial robustness with the distributional shift. However, the solutions they provided are either subject to additional supervisions, complicated causal graph and training strategies, or parallel to the existing adversarial training variants. Meanwhile, this paper provides a more feasible causal explanation for the adversarial vulnerability, by which we can design an effective plug-and-play causal regularization.

\noindent\textbf{Causal Graph and Intervention}. Pearl's graphical model~\cite{pearl2016causal} is adopted in this paper, where directed edges indicate the causality between node variables. The causal graph of the proposed CiiV framework is illustrated in Figure~\ref{fig:1.2} (b), where $R, C, X, Y$ indicate retinotopic sampling mask, confounding pattern, image, and prediction, respectively. $X\leftarrow C\rightarrow Y$ denotes that confounder $C$ is a common cause, affecting the distribution of both $X$ and $Y$, \eg, the edge in Figure~\ref{fig:1.1} (a). $X\to Y$ denotes the desired causality that a robust model is expected to learn. To achieve that, the ultimate goal of causal intervention is to identify the causal effect of $X\rightarrow Y$ by removing all spurious correlations~\cite{Judea2018thebookofwhy}, denoted as $P(Y|do(X=x))$. It can be either implemented as active intervention, like the randomized controlled trial, or passive $d$-separation~\cite{roy2021dseparation, pearl2016causal}, by which observing the confounder can block the spurious path, \eg, by conditioning on $C$, the dependency of path $X\leftarrow C\rightarrow Y$ is blocked.

\section{A Causal View on Adversarial Attack}
\label{sec.4}

In causality~\cite{pearl2016causal}, the total effect and causal effect of a predictive model based on the input $X$ can be defined as $P(Y|X)$, $P(Y|do(X=x))$, respectively. Given the proposed causal graph in Figure~\ref{fig:1.1}, the confounding path $X\leftarrow C\to Y$ causes the inequality between the above two, and thus the confounding effect can be represented as their difference.

Meanwhile, the general adversarial attack can be formulated as maximizing the probability of a tampered category $Y=\hat{y}$ within the budget $\mathcal{D}_\epsilon$~\cite{ren2020adversarial}, denoted as follows:
\begin{equation}
    \max_{\delta \in \mathcal{D}_\epsilon} \; P(Y=\hat{y}|X=x+\delta) \propto \sum_i \hat{y}_i log(p_i),
    \label{eq:4.1}
\end{equation}
where $\hat{y}=y' (y' \neq y)$ for targeted attack, $\hat{y}=-y$ for untargeted attack; $\hat{y}_i$ and $p_i$ are $i$-th entries of $\hat{y}$ and prediction $p$, respectively; $\delta$ is the additive perturbation; budget $\mathcal{D}_\epsilon$ is usually considered as an enclosing ball under $l_2 / l_\infty$ norm within radius $\epsilon$~\cite{zheng2019dist, kurakin2016adversarial}.

\begin{wrapfigure}{r}{8cm}
\vspace{-3mm}
\includegraphics[width=8cm]{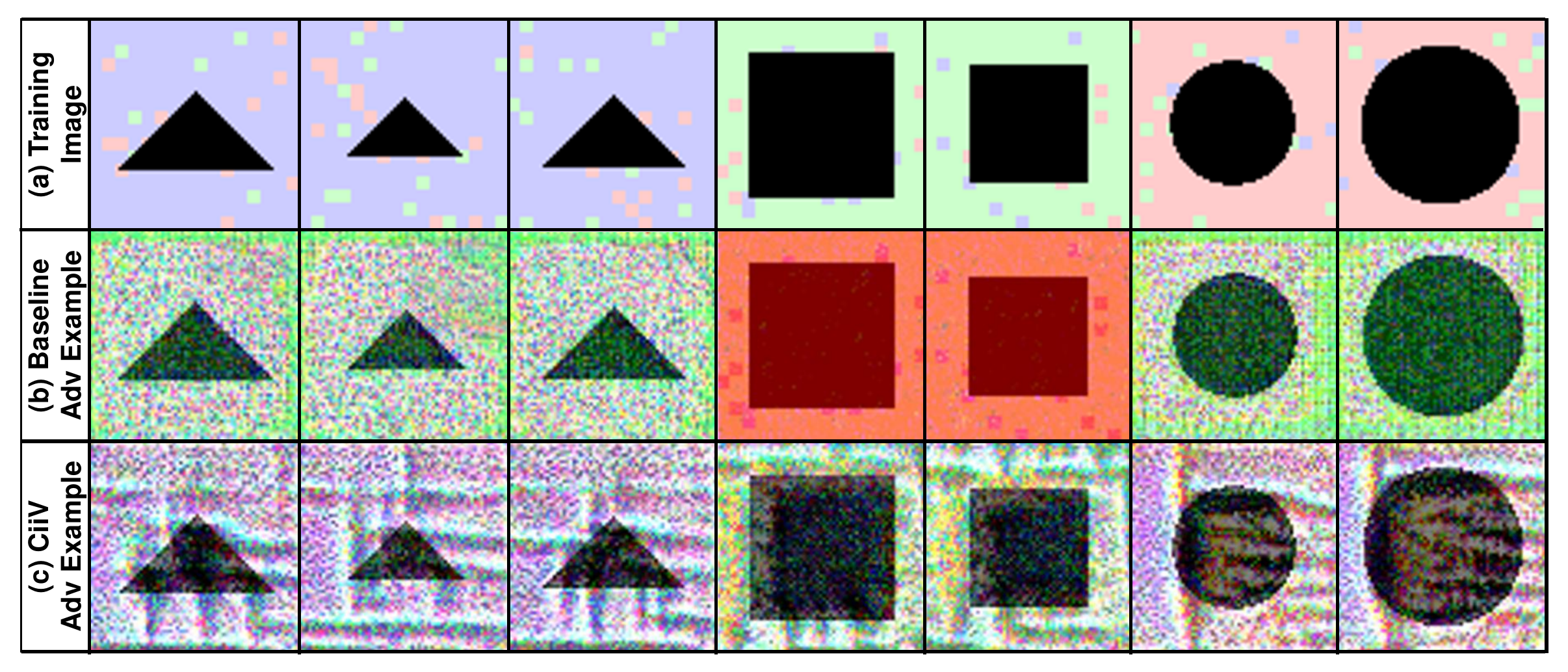}
\caption{(a) A Confounded-Toy Dataset with images that are composed of causal geometries and confounding color blocks. The adversarial examples generated by the model (c) w/ and (b) w/o the proposed CiiV.}
\label{fig:4}
\end{wrapfigure}

Notably, a valid $\mathcal{D}_\epsilon$ is not allowed to change the semantic structures, as they are designed to be imperceptible, \ie, the causal effect $P(Y|do(X=x))$ is invariant to $\delta$. Otherwise, the perturbation would become a ``poisoning'' that is beyond our scope~\cite{tramer2020fundamental}. Therefore, Eq
\ref{eq:4.1} essentially equals to maximize a tampered confounding effect through perturbations: $\max_{\delta} P(Y=\hat{y}|X=x+\delta) - P(Y=\hat{y}|do(X=x+\delta))$, \textit{subject to} $P(Y=\hat{y}|do(X=x+\delta)) = P(Y=\hat{y}|do(X=x))$, which applies to all kinds of attacks~\cite{gdfl2014explaining, madry2019deep, brendel2017decision, chen2017zoo, kurakin2016adversarial}.

To intuitively demonstrate the above causal theories, we design a Confounded-Toy dataset as shown in Figure~\ref{fig:4} (a), where images are composed of causal geometries and confounding color blocks. Similar to our example in Figure~\ref{fig:1.1}, a model directly trained on this dataset will recklessly learn the stochastic color block $C$ that shows statistical correlation with the category as the indicator of $Y$. As a result, adversarial examples generated by a PGD attacker on this model mainly tamper the confounding patterns (Figure~\ref{fig:4} (b)). In contrast, the proposed CiiV regularization forces the model to learn causal features instead, so it can only be fooled by poisoning the geometry (Figure~\ref{fig:4} (c)). More details of this Confounded-Toy dataset will be introduced in Appendix~\ref{sec:A}.

\section{A Causal View on Adversarial Defense}
\label{sec.5}

\begin{wrapfigure}{r}{6cm}
\vspace{-3mm}
\includegraphics[width=6cm]{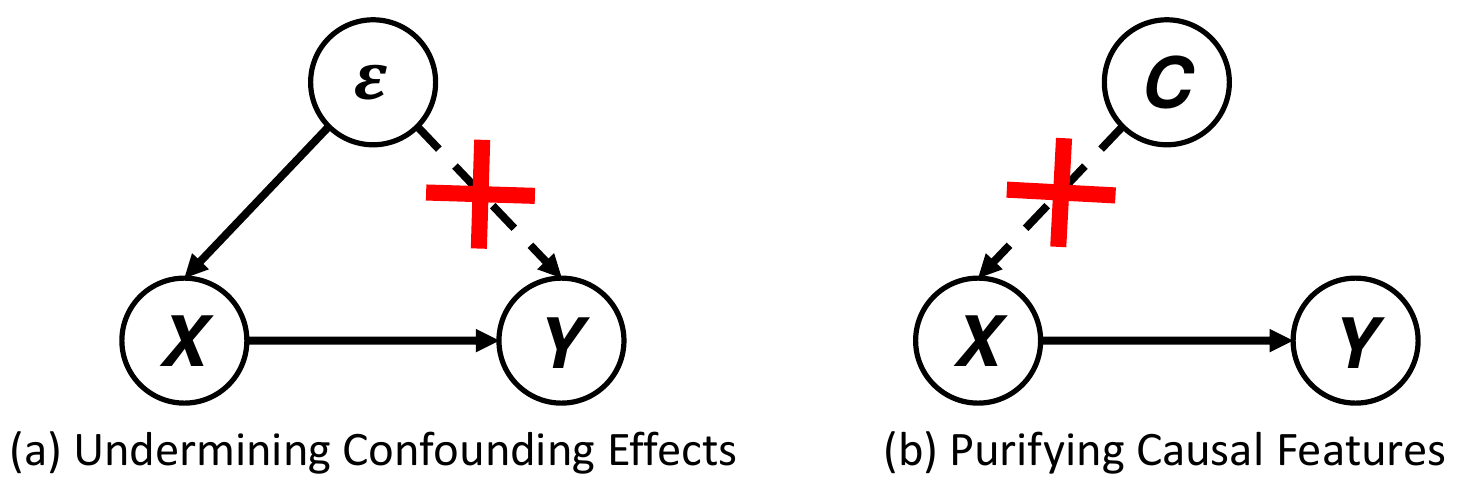}
\caption{Two common strategies to increase the adversarial robustness.}
\label{fig:5}
\vspace{-5mm}
\end{wrapfigure}

It has been acknowledged that directly adjusting an unknown confounder $C$ for $P(Y|do(X=x))$ without any assumption is impractical in causality field~\cite{d2019multi}. Due to the fact that adversarial examples are governed by unobserved confounding effects, most of the existing defending methods have to either intuitively assume a generative noise $\varepsilon$ to be the underline $C$ or assume $C$ to be certain identifiable noisy features that can be explicitly purified.

Specifically, adversarial training and its variants~\cite{szegedy2013intriguing, wong2019fast, dong2020api} together with some certified defenders like randomized smoothing~\cite{cohen2019certified} design some additive noises $\varepsilon$ to imitate adversarial perturbations, then undermine the confounding effect by asking the model to be robust against $\varepsilon$. On the other hand, the de-noising approaches, no matter the pre-network de-noising~\cite{xie2017mitigating, samangouei2018defense, buckman2018thermometer} or the in-network de-noising~\cite{li2020enhancing, xie2019feature} consider confounders to be explicitly removable patterns. Therefore, these common strategies can be summarized by two graphical operations as shown in Figure~\ref{fig:5}. 

However, we can neither guarantee the $C$ to be equal to $\varepsilon$, nor ensure all possible $C$ to be disentangled and purified. Relying on the  observation of such assumptive $C$ will at best make the above defenders robust against a subset of potential confounders. 

Among all existing defenders, mixup~\cite{zhang2018mixup} is most related to the proposed CiiV. It intervenes an image $x_i$ by linearly fusing with another image $x_j$, then forces the prediction $Y$ similar to the same combination of their one-hot labels. Yet, a valid instrumental variable is required to be independent of the confounder as we will introduce in the next section. Unfortunately, a new image $x_j$ can still depend on the same confounder of $x_i$. Recent studies~\cite{moosavi2017universal, zhang2020understanding} found that universal adversarial perturbations across images also exist, which explains why mixup cannot survive strong attackers.

\section{Approach}
\label{sec.6}

After connecting the adversarial vulnerability to the confounding effect learned by DNN models, the remaining question is how to obtain the pure causal effect, which is equivalent to applying causal intervention $P(Y|do(X=x))$ on the deep learning. Generally, there are four major interventions:  randomized controlled trial, backdoor adjustment, front-door adjustment, and instrumental variable estimation. However, the randomized controlled trial requires the control over causal features, the backdoor and front-door adjustments assume confounders or mediators to be observed, which are impractical for imperceptible adversarial perturbations. Therefore, we are interested in the last instrumental variable estimation that does not require such assumptions.

\subsection{Instrumental Variable Estimation}
\label{sec.6.1}
According to the definition~\cite{baiocchi2014instrumental, guo2016control}, a valid instrumental variable should satisfy: 1) it is independent of the confounder variable; 2) it affects $Y$ only through $X$. The instrumental variable can help to extract the causal effect of $X\to Y$ from $R\to X\to Y$, which is not confounded by $C$ ($d$-separated).

\begin{wrapfigure}{r}{7cm}
\vspace{-3mm}
\includegraphics[width=7cm]{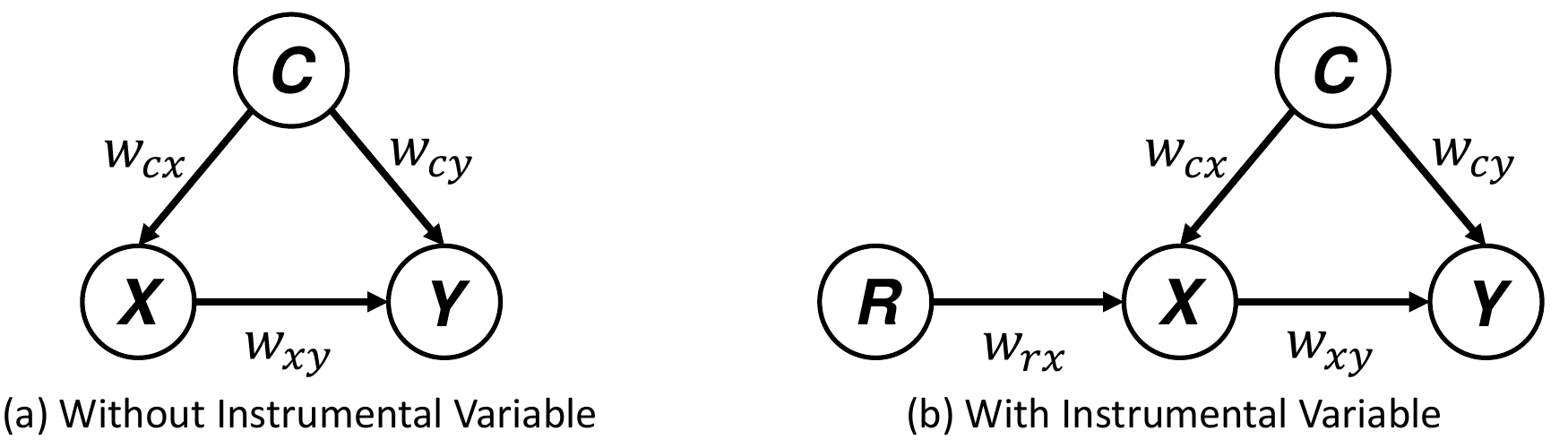}
\caption{The causal graphs w/ and w/o the instrumental variable. Nodes are assumed to be linked through linear associations $w_*$.}
\label{fig:6.1}
\vspace{-3mm}
\end{wrapfigure}

To better demonstrate the use of the instrumental variable~\cite{bowden1990instrumental}, we design two linear confounded models w/ and w/o the instrumental variable as shown in Figure~\ref{fig:6.1}. All variables are assumed to be linked by linear weights $w_*$. The confounder is an independent variable sampled from a normal distribution: $C \sim \mathcal{N}(0,1)$. The total effect and causal effect of $X\to Y$ can be represented as $Y[X=x]=w_{xy}x + w_{cy}c$ and $Y[do(X=x)]=w_{xy}x$, respectively. Note that we slightly abuse the notation of normalized effects $P(Y|X)$ and $P(Y|do(X=x))$, and use the form of unnormalized logits for simplicity. 

In the given confounded model of Figure~\ref{fig:6.1} (a), since $X$ is dependent on the confounder $C$ as $x=w_{cx}c + b_x$, where $b_x$ is the independent component of $X$, we cannot directly estimate the causal effect $Y[do(X=x)]$ by simply applying linear regression on $(x,y)$ pairs.

If $C$ is observable, the causal intervention could be conducted using the backdoor adjustment: $P(y|do(x)) = \sum_{c}P(y|x,c)P(c)$. The causal effect is thus estimated from the total effect by the observed $c$ and its distribution:
\begin{equation}
    Y[do(X=x)] = w_{xy}x + w_{cy}\sum_c c \cdot P(c) = w_{xy}x,
    \label{eq.3}
\end{equation}
where the confounding effect degrades to a constant as $\sum_c c \cdot P(c) = 0$.

However, if $C$ is unobservable, both backdoor adjustment and the causal graph in Figure~\ref{fig:6.1} (a) cannot remove the confounding effect. To this end, the instrumental variable $R$ is introduced as shown in Figure~\ref{fig:6.1} (b), where $X$ is now manipulated by both $C$ and $R$ as $x = w_{cx}c + w_{rx}r + b_x$. Due to the fact that $R$ is independent of $C$, the weight of causal link $X\to Y$ can be learned by applying different $r$ onto $(x,y)$ pairs, \ie, $y_{r_i} - y_{r_j} = w_{xy} (x_{r_i} - x_{r_j})$. The causal effect is thus estimated as follows:
\begin{equation}
    Y[do(X=x)] = \frac{y_{r_i} - y_{r_j}}{x_{r_i} - x_{r_j}} x = w_{xy}x,
\end{equation}
where subscripts $r_i\neq r_j$ indicate the value of $X$ and $Y$ under different instrumental variable $R$. The $d$-separated of $R\to X\leftarrow C$ ensures the subtraction eliminating the confounding effect during training. 

\subsection{The Proposed CiiV}
\label{sec.6.2}

With the help of instrumental variable $R$, the causal effect of the above linear example can be easily estimated. Yet, in practice, the effect of additive $R$ on an image is just as incomprehensible as the additive perturbation $C$, which doesn't introduce any useful inductive bias. Besides, the above subtraction is also hard to converge during backpropagation, as it may generate confusing gradients with opposite directions of $y_{r_i}$ and $y_{r_j}$. 

\begin{wrapfigure}{r}{7cm}
\vspace{-3mm}
\includegraphics[width=7cm]{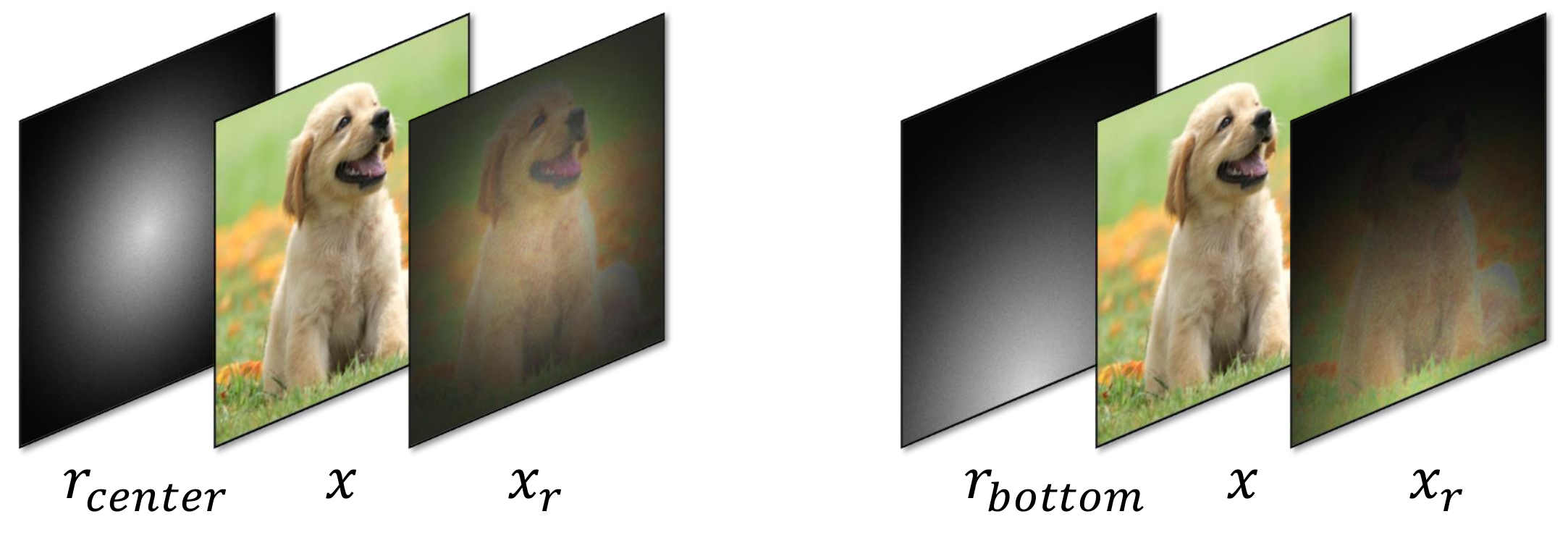}
\caption{Examples of retinotopic sampling and how it serves as the instrumental variable.}
\label{fig:6.2}
\end{wrapfigure}

In the proposed CiiV framework, we consider the retinotopic sampling mask as a multiplying instrumental variable and use it to augment the original dataset like Figure~\ref{fig:6.2}. Inspired by the human vision, the retina is known to consist of photoreceptors and a variety of other neurons~\cite{reddy2020biologically}. Retinotopic sampling is the result of the non-uniformly spatial distribution of these receptors~\cite{kolb1995webvision, arcaro2009retinotopic},  where the central fovea is significantly denser than the peripheral. It means that human vision is spatially lopsided by a centralized mask, which inspires us to adopt the retinotopic sampling mask with different centers as the instrumental variable $R$. Luckily, it also satisfies the requirements of a valid instrumental variable discussed in Section.~\ref{sec.6.1}: 1) the pre-defined retinotopic mask is guaranteed to be independent of any confounder in an image; 2) its effect on the prediction $Y$ can only pass through the change of causal features, as the non-robust confounders won't manifest stable patterns under different $R$. More detailed motivations behind the proposed CiiV will be discussed in Appendix~\ref{sec:B}.

Therefore, we adopt the multiplying retinotopic mask as our instrumental variable $R$ and design $R\to X$ to be an augmentation function on image $x_r=f(x, r)$, where $f(\cdot)$ applies different retinotopic sampling masks $r$ onto the confounded image $x$. The function is implemented as a differentiable multiplication layer and proved not to suffer from gradient obfuscation~\cite{athalye2018obfuscated} in Section~\ref{sec.7}. Detailed designs and experiments of $f(x, r)$ are investigated in Appendix~\ref{sec:C}.

Intuitively, when an object moves from the corner of our eyes to the center, the recognizability monotonously increases with the proportion of its captured contour, so we assume that the causal effect is linearly corresponding to the spatial coverage $\alpha_r$ of a retinotopic mask $r$ while the confounding effect is not. It is also consistent with previous findings~\cite{du2020rain} that visual confounders are usually high-frequency local components unevenly distributed in space. The relationship between the total effect and causal effect can thus be written as follows:
\begin{equation}
    Y[X=x_r] = w_{xy} x_r + w_{cy} c \approx  \alpha_r Y[do(X=x)] + w_{cy} c. 
\end{equation}
Note that we don't need to explicitly observe the above $c$. We can directly model the $Y[do(X=x)]$ by assigning different $r$ instead. The trick lies in the proposed CiiV regularization loss as follows:
\begin{equation}
    L_{CiiV} = \sum_{r_i\neq r_j} \Vert \alpha_{r_j} Y[X=x_{r_i}] - \alpha_{r_i} Y[X=x_{r_j}] \Vert,
    \label{eq:6.2}
\end{equation}
where $r_i$ and $r_j$ are two retinotopic sampling masks with spatial coverage $\alpha_{r_i}$ and $\alpha_{r_j}$, just like $r_{center}$ and $r_{bottom}$ in Figure~\ref{fig:6.2}. Since $w_{cy} c$ is independent of $r$, the above regularization can thus force the model to suppress the confounding effect. In practice, we implement CiiV as an $L_1$ loss on the feature space extracted by the backbone rather than the logit space, as the classifier weights can be taken out of the above regularization. Otherwise, the $L_{CiiV}$ could hurt the learning of the classifier. The overall training loss would be the combination of the conventional cross-entropy loss and the proposed CiiV loss with a trade-off parameter as $L_{All}=L_{CE} + \beta L_{CiiV}$.

\section{Experiments}
\label{sec.7}

\begin{table*}
\centering
\vspace{-2mm}
\scalebox{0.67}
{
\begin{tabular}{c ||c |c |c |c |c |c ||c |c |c |c |c |c}
\hline
\hline
Datasets & \multicolumn{6}{c||}{CIFAR-10} & \multicolumn{6}{c}{CIFAR-100} \\ 
\hline 
Attackers & Clean & FGSM & PGD-10 & AA-$l_{\infty}$ & AA-$l_2$  & Overall & Clean & FGSM & PGD-10 & AA-$l_{\infty}$ & AA-$l_2$ & Overall\\ 
\hline 
Baseline & 94.42 & 30.82 & 0.04 & 0.0 & 0.0 & 25.06 & 74.53 & 4.21 & 0.0 & 0.0 & 0.0 & 15.75 \\
mixup & \textbf{95.31} & 50.41 & 2.23 & 0.0 & 0.0 & 29.59 & \textbf{77.32} & 16.60 & 0.49 & 0.0 & 0.0 & 18.88 \\
BPFC & 90.21 & 24.58 & 6.19 & 2.92 & 35.55 & 31.89 & 61.48 & 17.00 & 10.23 & 7.17 & 29.16 & 25.01 \\
RS & 83.44 & 53.58 & 47.06 & 40.10 & 75.02 & 59.84 & 54.63 & 26.62 & 20.21 & 18.50 & 47.26 & 33.44 \\
(ours) CiiV & 86.89 & 64.44 & 50.75 & 43.23 & 82.48 & 65.56 & 58.88 & 32.48 & 23.63 & 23.05 & 55.40 & 38.69 \\
(ours) CiiV+mixup & 87.14 & 65.28 & 53.49 & \textbf{47.24} & 81.97 & 67.02 & 56.90 & 35.48 & \textbf{27.56} & \textbf{26.44} & 53.14 & 39.90 \\
(ours) CiiV+RandAug & 89.12 & \textbf{67.96} & \textbf{55.01} & 47.14 & \textbf{83.77} & \textbf{68.60} & 59.26 & \textbf{36.10} & 26.25 & 25.59 & \textbf{55.81} & \textbf{40.60} \\
\hline
AT$_{FGSM}$ & \textbf{84.52} & 54.42 & 43.84 & 37.94 & 60.20 & 56.18 & 51.99 & 26.27 & 22.54 & 18.31 & 31.06 & 30.03 \\
AT$_{PGD-10}$ & 83.94 & 52.90 & 47.19 & 43.18 & 55.46 & 56.53 & \textbf{56.48} & 25.99 & 22.56 & 20.04 & 28.96 & 30.81 \\
(ours) CiiV+AT$_{FGSM}$ & 83.67 & 67.28 & 57.96 & 50.93 & \textbf{80.09} & 67.99 & 53.83 & \textbf{39.00} & 32.20 & 30.48 & \textbf{50.47} & \textbf{41.20} \\
(ours) CiiV+AT$_{PGD-10}$ & 81.35 & \textbf{68.11} & \textbf{59.72} & \textbf{54.21} & 78.97 & \textbf{68.47} & 51.73 & 38.59 & \textbf{33.85} & \textbf{32.01} & 49.39 & 41.11 \\
\hline
\hline
\end{tabular}
}
\caption{The performances of white-box attack on CIFAR-10 and CIFAR-100. The upper half contains the AT-free defenders while the bottom half reports the AT-involved defenders.}
\label{tab:1}
\vspace{-5mm}
\end{table*}

\subsection{Datasets and Settings}
\label{sec.7.1}

\noindent\textbf{Datasets.} We evaluated the proposed CiiV and other defenders on three benchmark datasets: CIFAR-10, CIFAR-100, and mini-ImageNet~\cite{vinyals2016matching}. Both CIFAR-10 and CIFAR-100 contain 60K samples with the size of 32x32. mini-ImageNet is originally proposed by \cite{vinyals2016matching} for few-shot recognition, which consists of 100 classes and each has 600 images. We scaled the size of images to be 64x64 and split them into train/val/test sets with 42k/6k/12k images.

\begin{wraptable}{r}{8.5cm}
\centering
\vspace{-5mm}
\caption{The white-box attack on mini-ImageNet.}
\scalebox{0.67}
{
\begin{tabular}{c |c |c |c |c |c |c}
\hline
\hline 
Datasets & \multicolumn{6}{c}{mini-ImageNet} \\
\hline
Attackers & Clean & FGSM & PGD-10 & AA-$l_{\infty}$ & AA-$l_2$  & Overall\\ 
\hline 
Baseline & 71.17 & 1.37 & 0.01 & 0.0 & 0.0 & 14.51 \\
mixup & \textbf{73.88} & 2.96 & 0.0 & 0.0 & 0.0 & 15.37 \\
BPFC & 55.34 & 9.37 & 3.58 & 1.74 & 31.91 & 20.39 \\
RS & 52.15 & 15.09 & 13.25 & 6.93 & 45.82 & 26.65 \\
(ours) CiiV & 49.18 & 19.03 & 9.02 & 8.73 & 46.08 & 26.41 \\
(ours) CiiV+mixup & 48.83 & 23.93 & 15.12 & 11.45 & 45.47 & 28.96 \\
(ours) CiiV+RandAug & 51.65 & \textbf{32.22} & \textbf{24.82} & \textbf{18.87} & \textbf{48.47} & \textbf{35.21} \\
\hline
AT$_{FGSM}$ & 45.62 & 22.12 & 9.22 & 3.70 & 21.39 & 20.41 \\
AT$_{PGD-10}$ & \textbf{49.79} & 20.20 & 16.57 & 13.52 & 31.92 & 26.40 \\
(ours) CiiV+AT$_{FGSM}$ & 44.66 & 30.53 & 23.83 & 18.76 & \textbf{41.57} & \textbf{31.87} \\
(ours) CiiV+AT$_{PGD-10}$  & 42.85 & \textbf{31.72} & \textbf{25.46} & \textbf{19.30} & 39.72 & 31.81 \\
\hline
\hline
\end{tabular}
}
\label{tab:2}
\vspace{-2mm}
\end{wraptable}

\noindent\textbf{Training Details.} We followed \cite{pang2020bag}'s project to set all the hyper-parameters and architectures. All models were trained using the SGD optimizer with 0.9 momentum and 5e-4 weight decay. Experiments were conducted on GTX 2080ti GPUs with 128 batch size and 110 total epochs. The learning rate was started with 0.1 and updated by the factor of 0.1 at the following epochs $\{10, 100, 105\}$. The trade-off parameter $\beta$ was also initialized by 0.1 then multiplied by 10 at epochs $\{25,50,75\}$. Nine retinotopic centers were selected by using the $1/6$, $1/2$, and $5/6$ of width and height for each image. ResNet18~\cite{he2016deep} was utilized as the default backbone.

\noindent\textbf{Details of Threat Models.} We mainly evaluated the defenders on the clean images together with four threat models: FGSM~\cite{gdfl2014explaining}, PGD-10~\cite{madry2019deep}, AA-$l_{\infty}$ (AutoAttack $l_{\infty}$), and AA-$l_2$ (AutoAttack $l_2$)~\cite{croce2020reliable}. For FGSM and PGD-10, the adversarial perturbations were created under $l_\infty$ norm, where the budget radius $\epsilon$ was 8/255. PGD-10 ran 10 iterations with step size 2/255. AutoAttack is a recently released parameter-free attack that achieves the state-of-the-art attacking success rate under various defenders. It also prevents the model from gaining a false sense of security from the obfuscated gradients~\cite{athalye2018obfuscated}. We set the only parameter $\epsilon$ of AA-$l_{\infty}$ and AA-$l_2$ to be 8/255 and 0.5, respectively.

\noindent\textbf{Details of Defenders.} We divided the defenders into Adversarial Training (AT-involved) and AT-free approaches. For AT-involved, we adopted two popular defenders: AT$_{FGSM}$, AT$_{PGD-10}$, using the same parameters as the corresponding threat models. For AT-free methods, we investigated mixup~\cite{zhang2018mixup}, BPFC~\cite{addepalli2020towards}, and randomized smoothing(RS)~\cite{cohen2019certified}. The implementations of mixup and BPFC were directly adopted from their official github repositories. RS was re-implemented in our framework with $\sigma=0.25$ and the number of test trials $n=10$. The proposed CiiV itself is also an AT-free method. As a general regularization that is parallel to the above algorithms, we investigated its combination with other defenders as well.

\subsection{Diagnosis of Adversarial Robustness}
\label{sec.7.2}

The evaluation of adversarial robustness is always controversial as it can easily suffer from flawed or incomplete attack settings. To better eliminate the potential wrong sense of security, we followed \cite{carlini2019evaluating} to design our experiments and conduct a series of sanity checks at the end of this section.

\begin{wraptable}{r}{7.5cm}
\centering
\vspace{-5mm}
\caption{The performances of targeted PGD-10 under four different targeting settings: untargeted (UT), targeted by most likely / random / least likely categories (T-most, T-random, T-least).}
\scalebox{0.8}
{
\begin{tabular}{c |c |c |c |c}
\hline
\hline 
Datasets & \multicolumn{4}{c}{CIFAR-10} \\
\hline
Settings & UT & T-most & T-random & T-least\\ 
\hline 
CiiV & 50.75 & 55.62 & 71.05 & 74.37 \\
CiiV+mixup & 53.49 & 55.87 & 73.64 & 78.48 \\
CiiV+RandAug & 55.01 & 59.18 & 74.70 & 77.79 \\
CiiV+AT$_{FGSM}$ & 57.96 & 59.44 & 74.31 & 77.77 \\
CiiV+AT$_{PGD-10}$ & 59.72 & 60.46 & 75.21 & 78.42 \\
\hline
\hline
\end{tabular}
}
\label{tab:3}
\vspace{-2mm}
\end{wraptable}

\noindent\textbf{Adversarial Robustness Against White-box Attack.} As reported in Table~\ref{tab:1} and Table~\ref{tab:2}, we applied multiple white-box attacks on all three datasets. The proposed CiiV and its variants achieved better overall performances among both AT-free and AT-involved divisions. Note that Random Augmentation(RandAug)~\cite{cubuk2020randaugment} is not an adversarial defending method, whose overall performances on three datasets were just 24.38, 16.03, and 15.22, respectively. However, combining CiiV with RandAug worked as well as combining CiiV with AT methods, especially in the real-world mini-ImageNet. It proves that CiiV is indeed a proactive defender that doesn't rely on observing confounders. We also found that AT methods made the model significantly overfit the given attacker in all datasets. Besides, when replacing the training samples of CiiV with AT examples, \ie, CiiV+AT, the robustness came with the price of decreasing clean performances. However, augmenting CiiV with other AT-free methods like mixup and RandAug improved both clean and adversarial performances, which further supported our efforts to design a proactive AT-free defender.

\begin{figure}[t]
   \vspace{-2mm}
   \begin{minipage}[b]{1.0\linewidth}
   \centerline{\includegraphics[width=140mm]{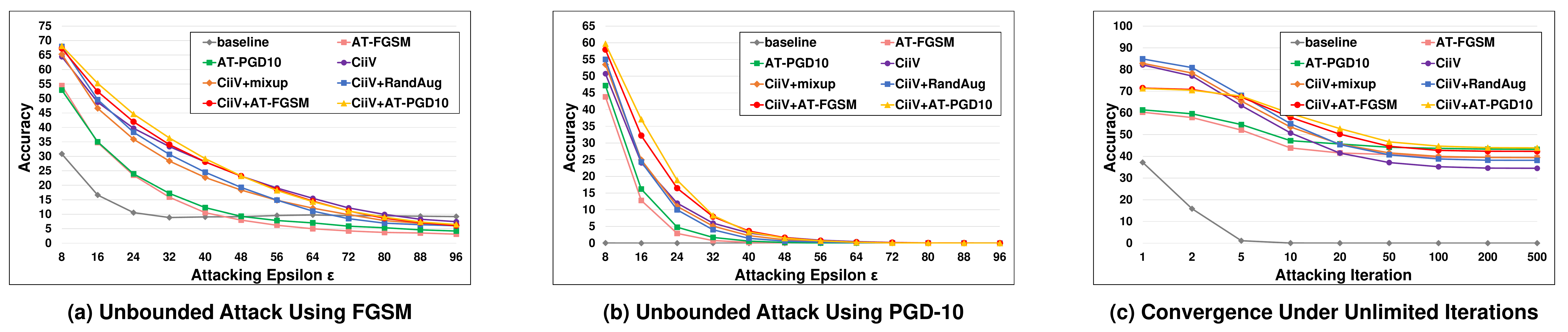}}
   \end{minipage}
   \caption{(a, b) Unbounded attacks on CIFAR-10 that increase the budget $\epsilon$ from 8/255 to 96/255. (c) The convergence of defenders under unlimited attacking iterations using PGD.}
   \label{fig:tab1}
   \vspace{-5mm}
\end{figure}

\noindent\textbf{Adversarial Robustness Against AutoAttack.} The state-of-the-art AutoAttack is an ensemble of diverse  parameter-free attacks~\cite{croce2020reliable}, including their proposed APGD-CE and APGD-DLR, the black-box Square Attack~\cite{andriushchenko2020square}, and the FAB attack~\cite{croce2020minimally} that is robust to obfuscated gradients~\cite{athalye2018obfuscated}. According to our experiments on AA-$l_{\infty}$ and AA-$l_2$, the proposed CiiV performed effectively on all of the above user-independent attacks. Moreover, combining CiiV with other defenders can further improve their adversarial robustness on both AA settings, proving that the CiiV is a general causal regularization parallel to most of the previous methods.

\noindent\textbf{Adversarial Robustness Against Targeted Attack.} The performances of the proposed CiiV under untargeted and targeted PGD-10 attacks were reported in Table~\ref{tab:3} using CIFAR-10 dataset. The targeted results were further divided into three protocols: 1) most likely category, 2) least likely category, and 3) random category. The results revealed that the confounding effect could also be the cause of ambiguous prediction, as similar categories are easier to be attacked. We also noticed that the performances under untargeted attack would be closer to the most likely targeted attack when the robustness of the model increases. It's probably because the similar categories share the similar confounder distributions, \ie, environments, and thus utilized by the attacker.

\noindent\textbf{Adversarial Robustness Under Unbounded Attack.}
To evaluate the validity of defenders, we demonstrated the performances of CiiV and its variants together with the baseline and two AT models under unbounded attacking in Figure~\ref{fig:tab1} (a, b). When the budget $\epsilon$ of the attacker was increased from 8/255 to 96/255, all performances were either converged to 0\% accuracy for the strong PGD attack or converged to the random guesses for the weak FGSM attack. Any valid defender shouldn't survive such an unbounded attack, as it allows the attacker to modify the entire image and erasing all causal features. We also tested unlimited iterations of PGD attack, all CiiV and its variants are successfully converged after 100 iterations as shown in Figure~\ref{fig:tab1} (c). Note that AT and CiiV+AT are more robust than other defenders in this setting, which is probably caused by the exposure of adversarial examples during training.

\noindent\textbf{Ablation Studies.} In this paragraph, we evaluated the effectiveness of different settings and parameters of the proposed CiiV. As reported in Table~\ref{tab:4}, 1) we investigated the $L_1$ and $L_2$ versions of the CiiV loss, where the $L_1$ is slightly better than $L_2$; 2) we tried random assignments of the retinotopic centers as R=$\{r_{rand}\}$, which is very close to our fixed centers; 3) we also reported the performances of retinotopic augmentation only as RetiAug, which had higher clean performance but worse adversarial robustness than CiiV. Note that RetiAug itself can also be treated as an approximation of CiiV by assigning all $\alpha$ to 1.0. Besides, cross-entropy losses under different $r$ also forced the model to ignore the non-robust confounding patterns; 4) other choices of hyper-parameters of CiiV were also reported, we found that $\beta$ empirically served as a trade-off between clean performance and adversarial robustness, and applying more retinotopic sampling masks (larger $N_R$) would make a better estimation, yet, its improvements got converged. Additional studies and experiments will be given in Appendix.

\begin{wraptable}{r}{8.5cm}
\centering
\vspace{-5mm}
\caption{Ablation Studies of CiiV on CIFAR-100.}
\scalebox{0.75}
{
\begin{tabular}{c |c |c |c |c |c |c}
\hline
\hline 
Datasets & \multicolumn{6}{c}{CIFAR-100} \\
\hline
Attackers & Clean & FGSM & PGD-10 & AA-$l_{\infty}$ & AA-$l_2$  & Overall\\ 
\hline
$L_1$ CiiV & 58.88 & 32.48 & 23.63 & 23.05 & 55.40 & 38.69 \\
$L_2$ CiiV & 57.93 & 31.78 & 22.27 & 22.28 & 54.51 & 37.75 \\
R=$\{r_{rand}\}$ & 58.79 & 32.20 & 22.35 & 22.90 & 55.28 & 38.30 \\
RetiAug & 61.88 & 31.69 & 20.29 & 18.32 & 53.19 & 37.07 \\
$\beta = 0.01$ & 59.85 & 32.00 & 21.52 & 20.85 & 54.23 & 37.69 \\
$\beta = 1.0$ & 55.26 & 34.48 & 26.80 & 25.02 & 51.82 & 38.68 \\
$N_R=2$ & 56.45 & 30.47 & 21.57 & 21.03 & 53.39 & 36.58 \\
$N_R=5$ & 58.34 & 32.01 & 22.34 & 22.61 & 54.58 & 37.98 \\

\hline
\hline
\end{tabular}
}
\label{tab:4}
\vspace{-2mm}
\end{wraptable}

\begin{figure}[t]
   \begin{minipage}[b]{1.0\linewidth}
   \centerline{\includegraphics[width=140mm]{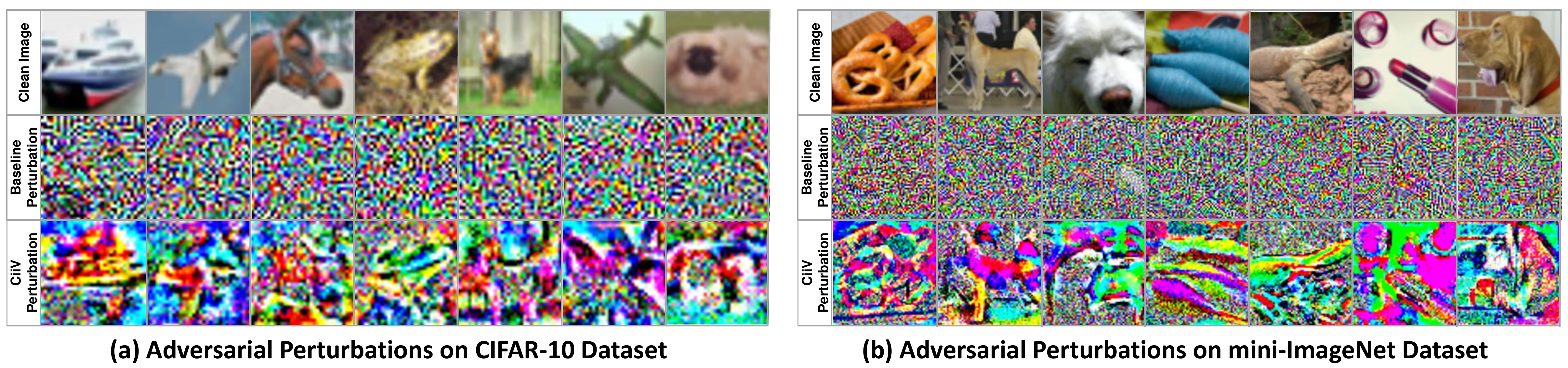}}
   \end{minipage}
   \caption{Generated perturbations of models w/ and w/o CiiV on CIFAR-10 and mini-ImageNet.}
   \label{fig:7}
   \vspace{-5mm}
\end{figure}

\noindent\textbf{Visualization.} We visualized the generated PGD perturbations for models w/ and w/o CiiV in Figure~\ref{fig:7}. It demonstrates that the baseline models can be easily fooled by imperceptible confounders while the proposed CiiV forces the model to learn causal features, as the adversarial perturbations have to erase the structural patterns to fool the CiiV model.

\noindent\textbf{The Evaluation Checklist.} To verify that the proposed CiiV doesn't suffer from flawed or incomplete evaluations, the above experiments were designed to follow a series of sanity checks introduced by \cite{carlini2019evaluating}: 1) Iterative attacks are better than single-step attacks, \eg, PGD \textit{vs} FGSM in Table~\ref{tab:1}\&\ref{tab:2} and Figure~\ref{fig:tab1}.
2) Unbounded adversarial examples become random guessing or 0\% accuracy, \eg, Figure~\ref{fig:tab1} (a,b). 
3) The accuracy converges with the increasing of attack steps: Figure~\ref{fig:tab1} (c).
4) Investigating both targeted attacks and untargeted attacks, \eg, Table~\ref{tab:3}.
5) Using black-box attacks and the attacks circumventing obfuscated gradients to avoid the potentially flawed adversarial example generation, \eg, the results under AA-$l_{\infty}$ and AA-$l_2$ in Table~\ref{tab:1}\&\ref{tab:2}. 

\section{Conclusion}
\label{sec.8}

In this paper, we presented a CiiV defender that worked as a general causal regularization without the need for adversarial examples. CiiV consists of a spatial data augmentation using different retinotopic sampling masks, and a regularization loss that encourages the model to suppress local confounding patterns by learning features linearly responding to spatial interpolations. We followed the checklist from \cite{carlini2019evaluating} to design our evaluation experiments and adopted the user-independent AutoAttack~\cite{croce2020reliable} as the main indicator of adversarial robustness. Extensive experiments on all settings proved that CiiV is robust against various adaptive attacks, and it can also serve as a plug-and-play regularization for other defenders. Besides, this paper also provides a fundamental viewpoint of the relationship between adversarial robustness and causal intervention, which may guide the design of future defenders

\section*{Ethics Statement}

Due to the fact that deep learning algorithms have been widely deployed in the present recommendation system, person identification system, and automatic/assist driving system, the potential ethical problems are also growing. The adversarial robustness field studied by this paper is one of the core efforts to address these concerns. Without the adversarial robustness, the DNN-based computer vision systems are vulnerable to imperceptible noises, which threats the safety of human life and property. Specifically, recent studies of  physical-based attacks have proved that simply wearing specially designed clothes can fool an artificial recognition system. The increasing traffic accident caused by driver-assistance systems further confirms the above concerns. Therefore, we provide a general causal regularization that could easily be plugged into most of the current adversarial defending methods to further boost the robustness of the system, which may significantly reduce the chance of failure recognition caused by adversarial perturbations.

\section*{Reproducibility Statement}

In the past decades, the open-source movement of the machine learning community has greatly promoted the development of related fields. To ensure the reproducibility of the proposed method, we are going to publish our codes on the github. The anonymous version of our entire project is also available in supplementary materials. More detailed instructions and explanations of our codes will be added to the project before releasing to the public. All datasets used in this paper are also publicly available and can be easily found via torchvision package or other public github repositories.

\bibliographystyle{unsrt}  
\bibliography{references}

\begin{thebibliography}{10}

\bibitem{gdfl2014explaining}
Ian~J Goodfellow, Jonathon Shlens, and Christian Szegedy.
\newblock Explaining and harnessing adversarial examples.
\newblock In {\em ICML}, 2015.

\bibitem{szegedy2013intriguing}
Christian Szegedy, Wojciech Zaremba, Ilya Sutskever, Joan Bruna, Dumitru Erhan,
  Ian Goodfellow, and Rob Fergus.
\newblock Intriguing properties of neural networks.
\newblock {\em ICLR}, 2013.

\bibitem{croce2020reliable}
Francesco Croce and Matthias Hein.
\newblock Reliable evaluation of adversarial robustness with an ensemble of
  diverse parameter-free attacks.
\newblock In {\em ICML}. PMLR, 2020.

\bibitem{tramer2020adaptive}
Florian Tramer, Nicholas Carlini, Wieland Brendel, and Aleksander Madry.
\newblock On adaptive attacks to adversarial example defenses.
\newblock {\em arxiv}, 2020.

\bibitem{athalye2018obfuscated}
Anish Athalye, Nicholas Carlini, and David Wagner.
\newblock Obfuscated gradients give a false sense of security: Circumventing
  defenses to adversarial examples.
\newblock In {\em International Conference on Machine Learning}, 2018.

\bibitem{kannan2018adversarial}
Harini Kannan, Alexey Kurakin, and Ian Goodfellow.
\newblock Adversarial logit pairing.
\newblock {\em arxiv}, 2018.

\bibitem{cui2020learnable}
Jiequan Cui, Shu Liu, Liwei Wang, and Jiaya Jia.
\newblock Learnable boundary guided adversarial training.
\newblock {\em arXiv preprint arXiv:2011.11164}, 2020.

\bibitem{tramer2017ensemble}
Florian Tram{\`e}r, Alexey Kurakin, Nicolas Papernot, Ian Goodfellow, Dan
  Boneh, and Patrick McDaniel.
\newblock Ensemble adversarial training: Attacks and defenses.
\newblock {\em ICLR}, 2018.

\bibitem{schott2018towards}
Lukas Schott, Jonas Rauber, Matthias Bethge, and Wieland Brendel.
\newblock Towards the first adversarially robust neural network model on mnist.
\newblock In {\em ICLR}, 2019.

\bibitem{zhang2019limitations}
Huan Zhang, Hongge Chen, Zhao Song, Duane Boning, Inderjit~S Dhillon, and
  Cho-Jui Hsieh.
\newblock The limitations of adversarial training and the blind-spot attack.
\newblock In {\em ICLR}, 2019.

\bibitem{gilmer2018adversarial}
Justin Gilmer, Luke Metz, Fartash Faghri, Samuel~S Schoenholz, Maithra Raghu,
  Martin Wattenberg, and Ian Goodfellow.
\newblock Adversarial spheres.
\newblock {\em Workshop of ICLR}, 2018.

\bibitem{schmidt2018adversarially}
Ludwig Schmidt, Shibani Santurkar, Dimitris Tsipras, Kunal Talwar, and
  Aleksander Madry.
\newblock Adversarially robust generalization requires more data.
\newblock {\em NeurIPS}, 2018.

\bibitem{xie2020adversarial}
Cihang Xie, Mingxing Tan, Boqing Gong, Jiang Wang, Alan~L Yuille, and Quoc~V
  Le.
\newblock Adversarial examples improve image recognition.
\newblock In {\em CVPR}, 2020.

\bibitem{ilyas2019adversarial}
Andrew Ilyas, Shibani Santurkar, Logan Engstrom, Brandon Tran, and Aleksander
  Madry.
\newblock Adversarial examples are not bugs, they are features.
\newblock {\em NeurIPS}, 2019.

\bibitem{salman2020adversarially}
Hadi Salman, Andrew Ilyas, Logan Engstrom, Ashish Kapoor, and Aleksander Madry.
\newblock Do adversarially robust imagenet models transfer better?
\newblock {\em arXiv preprint arXiv:2007.08489}, 2020.

\bibitem{herculano2012remarkable}
Suzana Herculano-Houzel.
\newblock The remarkable, yet not extraordinary, human brain as a scaled-up
  primate brain and its associated cost.
\newblock {\em Proceedings of the National Academy of Sciences}, 2012.

\bibitem{masland2001fundamental}
Richard~H Masland.
\newblock The fundamental plan of the retina.
\newblock {\em Nature Neuroscience}, 2001.

\bibitem{kim2020modeling}
Edward Kim, Jocelyn Rego, Yijing Watkins, and Garrett~T Kenyon.
\newblock Modeling biological immunity to adversarial examples.
\newblock In {\em CVPR}, 2020.

\bibitem{pearl2009causality}
Judea Pearl.
\newblock {\em Causality}.
\newblock Cambridge university press, 2009.

\bibitem{arcaro2009retinotopic}
Michael~J Arcaro, Stephanie~A McMains, Benjamin~D Singer, and Sabine Kastner.
\newblock Retinotopic organization of human ventral visual cortex.
\newblock {\em Journal of neuroscience}, 2009.

\bibitem{greenland2000introduction}
Sander Greenland.
\newblock An introduction to instrumental variables for epidemiologists.
\newblock {\em International journal of epidemiology}, 2000.

\bibitem{carlini2019evaluating}
Nicholas Carlini, Anish Athalye, Nicolas Papernot, Wieland Brendel, Jonas
  Rauber, Dimitris Tsipras, Ian Goodfellow, Aleksander Madry, and Alexey
  Kurakin.
\newblock On evaluating adversarial robustness.
\newblock {\em arXiv}, 2019.

\bibitem{wang2021understanding}
He~Wang, Feixiang He, Zhexi Peng, Yong-Liang Yang, Tianjia Shao, Kun Zhou, and
  David Hogg.
\newblock Understanding the robustness of skeleton-based action recognition
  under adversarial attack.
\newblock In {\em CVPR}, 2021.

\bibitem{qi2021stabilized}
Gege Qi, Lijun Gong, Yibing Song, Kai Ma, and Yefeng Zheng.
\newblock Stabilized medical image attacks.
\newblock In {\em ICLR}, 2021.

\bibitem{finlayson2019adversarial}
Samuel~G Finlayson, John~D Bowers, Joichi Ito, Jonathan~L Zittrain, Andrew~L
  Beam, and Isaac~S Kohane.
\newblock Adversarial attacks on medical machine learning.
\newblock {\em Science}, 2019.

\bibitem{xie2017adversarial}
Cihang Xie, Jianyu Wang, Zhishuai Zhang, Yuyin Zhou, Lingxi Xie, and Alan
  Yuille.
\newblock Adversarial examples for semantic segmentation and object detection.
\newblock In {\em ICCV}, 2017.

\bibitem{xiang2019generating}
Chong Xiang, Charles~R Qi, and Bo~Li.
\newblock Generating 3d adversarial point clouds.
\newblock In {\em CVPR}, 2019.

\bibitem{cisse2017houdini}
Moustapha Cisse, Yossi Adi, Natalia Neverova, and Joseph Keshet.
\newblock Houdini: Fooling deep structured prediction models.
\newblock {\em arXiv}, 2017.

\bibitem{carlini2018audio}
Nicholas Carlini and David Wagner.
\newblock Audio adversarial examples: Targeted attacks on speech-to-text.
\newblock In {\em Security and Privacy Workshops (SPW)}, 2018.

\bibitem{huang2017adversarial}
Sandy Huang, Nicolas Papernot, Ian Goodfellow, Yan Duan, and Pieter Abbeel.
\newblock Adversarial attacks on neural network policies.
\newblock {\em arXiv}, 2017.

\bibitem{diao2021basarblackbox}
Yunfeng Diao, Tianjia Shao, Yong-Liang Yang, Kun Zhou, and He~Wang.
\newblock Basar:black-box attack on skeletal action recognition.
\newblock In {\em CVPR}, 2021.

\bibitem{moosavi2016deepfool}
Seyed-Mohsen Moosavi-Dezfooli, Alhussein Fawzi, and Pascal Frossard.
\newblock Deepfool: a simple and accurate method to fool deep neural networks.
\newblock In {\em CVPR}, 2016.

\bibitem{kurakin2016adversarial}
Alexey Kurakin, Ian Goodfellow, Samy Bengio, et~al.
\newblock Adversarial examples in the physical world, 2016.

\bibitem{papernot2016limitations}
Nicolas Papernot, Patrick McDaniel, Somesh Jha, Matt Fredrikson, Z~Berkay
  Celik, and Ananthram Swami.
\newblock The limitations of deep learning in adversarial settings.
\newblock In {\em EuroS\&P}, 2016.

\bibitem{carlini2017towards}
Nicholas Carlini and David Wagner.
\newblock Towards evaluating the robustness of neural networks.
\newblock In {\em Symposium on Security and Privacy (SP)}. IEEE, 2017.

\bibitem{zheng2019dist}
Tianhang Zheng, Changyou Chen, and Kui Ren.
\newblock Distributionally adversarial attack.
\newblock In {\em AAAI}, 2019.

\bibitem{wong2019fast}
Eric Wong, Leslie Rice, and J~Zico Kolter.
\newblock Fast is better than free: Revisiting adversarial training.
\newblock In {\em ICLR}, 2019.

\bibitem{dong2020api}
Xinshuai Dong, Hong Liu, Rongrong Ji, Liujuan Cao, Qixiang Ye, Jianzhuang Liu,
  and Qi~Tian.
\newblock Api-net: Robust generative classifier via a single discriminator.
\newblock In {\em ECCV}, 2020.

\bibitem{zhang2018mixup}
Hongyi Zhang, Moustapha Cisse, Yann~N Dauphin, and David Lopez-Paz.
\newblock mixup: Beyond empirical risk minimization.
\newblock {\em ICLR}, 2018.

\bibitem{buckman2018thermometer}
Jacob Buckman, Aurko Roy, Colin Raffel, and Ian Goodfellow.
\newblock Thermometer encoding: One hot way to resist adversarial examples.
\newblock In {\em ICLR}, 2018.

\bibitem{xie2019feature}
Cihang Xie, Yuxin Wu, Laurens van~der Maaten, Alan~L Yuille, and Kaiming He.
\newblock Feature denoising for improving adversarial robustness.
\newblock In {\em CVPR}, 2019.

\bibitem{cohen2019certified}
Jeremy Cohen, Elan Rosenfeld, and Zico Kolter.
\newblock Certified adversarial robustness via randomized smoothing.
\newblock In {\em ICML}, 2019.

\bibitem{zhang2020causal}
Cheng Zhang, Kun Zhang, and Yingzhen Li.
\newblock A causal view on robustness of neural networks.
\newblock {\em NeurIPS}, 2020.

\bibitem{zhang2021adversarial}
Yonggang Zhang, Mingming Gong, Tongliang Liu, Gang Niu, Xinmei Tian, Bo~Han,
  Bernhard Sch{\"o}lkopf, and Kun Zhang.
\newblock Adversarial robustness through the lens of causality.
\newblock {\em arXiv preprint arXiv:2106.06196}, 2021.

\bibitem{yang2019causal}
Chao-Han~Huck Yang, Yi-Chieh Liu, Pin-Yu Chen, Xiaoli Ma, and Yi-Chang~James
  Tsai.
\newblock When causal intervention meets adversarial examples and image masking
  for deep neural networks.
\newblock In {\em ICIP}. IEEE, 2019.

\bibitem{singh2021learning}
Harvineet Singh, Shalmali Joshi, Finale Doshi-Velez, and Himabindu Lakkaraju.
\newblock Learning under adversarial and interventional shifts.
\newblock {\em arXiv preprint arXiv:2103.15933}, 2021.

\bibitem{pearl2016causal}
Judea Pearl, Madelyn Glymour, and Nicholas~P Jewell.
\newblock {\em Causal inference in statistics: A primer}.
\newblock John Wiley \& Sons, 2016.

\bibitem{Judea2018thebookofwhy}
Judea Pearl and Dana Mackenzie.
\newblock {\em The Book of Why: The New Science of Cause and Effect}.
\newblock Basic Books, 2018.

\bibitem{roy2021dseparation}
Jason~A. Roy, 2020.

\bibitem{ren2020adversarial}
Kui Ren, Tianhang Zheng, Zhan Qin, and Xue Liu.
\newblock Adversarial attacks and defenses in deep learning.
\newblock {\em Engineering}, 2020.

\bibitem{tramer2020fundamental}
Florian Tramer, Jens Behrmann, Nicholas Carlini, Nicolas Papernot, and
  Jorn-Henrik Jacobsen.
\newblock Fundamental tradeoffs between invariance and sensitivity to
  adversarial perturbations.
\newblock In {\em ICML}, 2020.

\bibitem{madry2019deep}
Aleksander Madry, Aleksandar Makelov, Ludwig Schmidt, Dimitris Tsipras, and
  Adrian Vladu.
\newblock Towards deep learning models resistant to adversarial attacks, 2018.

\bibitem{brendel2017decision}
Wieland Brendel, Jonas Rauber, and Matthias Bethge.
\newblock Decision-based adversarial attacks: Reliable attacks against
  black-box machine learning models.
\newblock {\em ICLR}, 2018.

\bibitem{chen2017zoo}
Pin-Yu Chen, Huan Zhang, Yash Sharma, Jinfeng Yi, and Cho-Jui Hsieh.
\newblock Zoo: Zeroth order optimization based black-box attacks to deep neural
  networks without training substitute models.
\newblock In {\em Workshop on artificial intelligence and security}, 2017.

\bibitem{d2019multi}
Alexander D'Amour.
\newblock On multi-cause causal inference with unobserved confounding:
  Counterexamples, impossibility, and alternatives.
\newblock In {\em AISTATS}, 2019.

\bibitem{xie2017mitigating}
Cihang Xie, Jianyu Wang, Zhishuai Zhang, Zhou Ren, and Alan Yuille.
\newblock Mitigating adversarial effects through randomization.
\newblock {\em ICLR}, 2018.

\bibitem{samangouei2018defense}
Pouya Samangouei, Maya Kabkab, and Rama Chellappa.
\newblock Defense-gan: Protecting classifiers against adversarial attacks using
  generative models.
\newblock {\em ICLR}, 2018.

\bibitem{li2020enhancing}
Guanlin Li, Shuya Ding, Jun Luo, and Chang Liu.
\newblock Enhancing intrinsic adversarial robustness via feature pyramid
  decoder.
\newblock In {\em CVPR}, 2020.

\bibitem{moosavi2017universal}
Seyed-Mohsen Moosavi-Dezfooli, Alhussein Fawzi, Omar Fawzi, and Pascal
  Frossard.
\newblock Universal adversarial perturbations.
\newblock In {\em CVPR}, 2017.

\bibitem{zhang2020understanding}
Chaoning Zhang, Philipp Benz, Tooba Imtiaz, and In~So Kweon.
\newblock Understanding adversarial examples from the mutual influence of
  images and perturbations.
\newblock In {\em CVPR}, 2020.

\bibitem{baiocchi2014instrumental}
Michael Baiocchi, Jing Cheng, and Dylan~S Small.
\newblock Instrumental variable methods for causal inference.
\newblock {\em Statistics in medicine}, 2014.

\bibitem{guo2016control}
Zijian Guo and Dylan~S Small.
\newblock Control function instrumental variable estimation of nonlinear causal
  effect models.
\newblock {\em The Journal of Machine Learning Research}, 2016.

\bibitem{bowden1990instrumental}
Roger~J Bowden and Darrell~A Turkington.
\newblock {\em Instrumental variables}.
\newblock Cambridge university press, 1990.

\bibitem{reddy2020biologically}
Manish~Vuyyuru Reddy, Andrzej Banburski, Nishka Pant, and Tomaso Poggio.
\newblock Biologically inspired mechanisms for adversarial robustness.
\newblock {\em NeurIPS}, 2020.

\bibitem{kolb1995webvision}
Helga Kolb, Eduardo Fernandez, and Ralph Nelson.
\newblock Webvision: the organization of the retina and visual system.
\newblock {\em book}, 1995.

\bibitem{du2020rain}
Jiawei Du, Hanshu Yan, Vincent~YF Tan, Joey~Tianyi Zhou, Rick Siow~Mong Goh,
  and Jiashi Feng.
\newblock Rain: A simple approach for robust and accurate image classification
  networks.
\newblock {\em arXiv preprint arXiv:2004.14798}, 2020.

\bibitem{vinyals2016matching}
Oriol Vinyals, Charles Blundell, Timothy Lillicrap, Daan Wierstra, et~al.
\newblock Matching networks for one shot learning.
\newblock {\em NeurIPS}, 2016.

\bibitem{pang2020bag}
Tianyu Pang, Xiao Yang, Yinpeng Dong, Hang Su, and Jun Zhu.
\newblock Bag of tricks for adversarial training.
\newblock {\em ICLR}, 2021.

\bibitem{he2016deep}
Kaiming He, Xiangyu Zhang, Shaoqing Ren, and Jian Sun.
\newblock Deep residual learning for image recognition.
\newblock In {\em CVPR}, 2016.

\bibitem{addepalli2020towards}
Sravanti Addepalli, Arya Baburaj, Gaurang Sriramanan, and R~Venkatesh Babu.
\newblock Towards achieving adversarial robustness by enforcing feature
  consistency across bit planes.
\newblock In {\em CVPR}, 2020.

\bibitem{cubuk2020randaugment}
Ekin~D Cubuk, Barret Zoph, Jonathon Shlens, and Quoc~V Le.
\newblock Randaugment: Practical automated data augmentation with a reduced
  search space.
\newblock In {\em CVPR Workshops}, 2020.

\bibitem{andriushchenko2020square}
Maksym Andriushchenko, Francesco Croce, Nicolas Flammarion, and Matthias Hein.
\newblock Square attack: a query-efficient black-box adversarial attack via
  random search.
\newblock In {\em ECCV}, 2020.

\bibitem{croce2020minimally}
Francesco Croce and Matthias Hein.
\newblock Minimally distorted adversarial examples with a fast adaptive
  boundary attack.
\newblock In {\em ICML}, 2020.

\bibitem{yang2021causalvae}
Mengyue Yang, Furui Liu, Zhitang Chen, Xinwei Shen, Jianye Hao, and Jun Wang.
\newblock Causalvae: disentangled representation learning via neural structural
  causal models.
\newblock In {\em CVPR}, 2021.

\bibitem{higgins2016beta}
Irina Higgins, Loic Matthey, Arka Pal, Christopher Burgess, Xavier Glorot,
  Matthew Botvinick, Shakir Mohamed, and Alexander Lerchner.
\newblock beta-vae: Learning basic visual concepts with a constrained
  variational framework.
\newblock 2017.

\bibitem{land2019eye}
Michael Land.
\newblock Eye movements in man and other animals.
\newblock {\em Vision research}, 2019.

\bibitem{uesato2018adversarial}
Jonathan Uesato, Brendan O’donoghue, Pushmeet Kohli, and Aaron Oord.
\newblock Adversarial risk and the dangers of evaluating against weak attacks.
\newblock In {\em ICML}, 2018.

\bibitem{simonyan2014very}
Karen Simonyan and Andrew Zisserman.
\newblock Very deep convolutional networks for large-scale image recognition.
\newblock {\em ICLR}, 2015.

\bibitem{zagoruyko2016wide}
Sergey Zagoruyko and Nikos Komodakis.
\newblock Wide residual networks.
\newblock {\em arXiv preprint arXiv:1605.07146}, 2016.

\end{thebibliography}

\clearpage

\appendix

\section{Details of the Confounded-Toy Dataset}
\label{sec:A}

\begin{wrapfigure}{r}{7.5cm}
\vspace{-3mm}
\includegraphics[width=7.5cm]{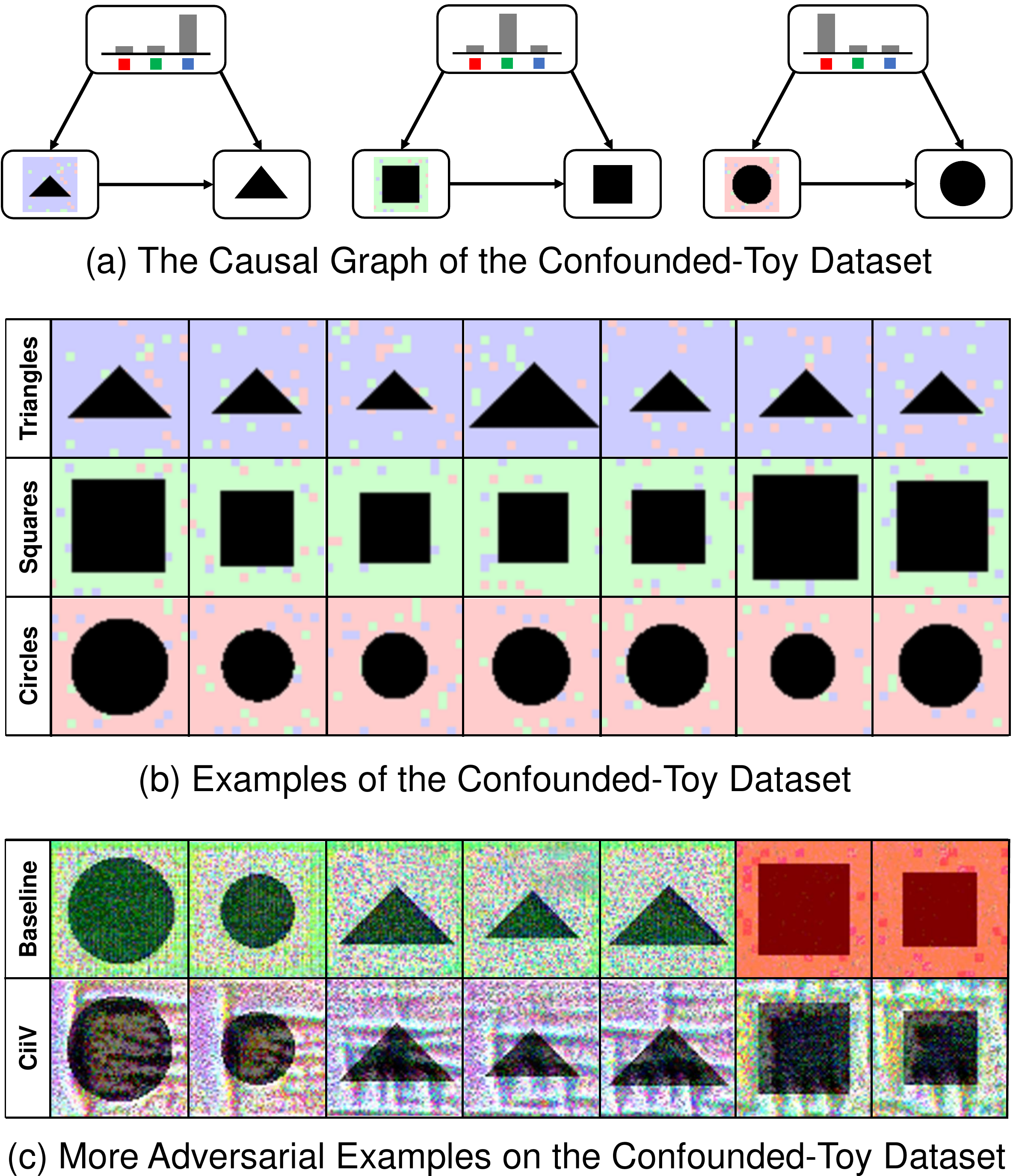}
\caption{(a) The causal graph of the Confounded-Toy dataset. (b) More examples of the proposed Confounded-Toy dataset. (c) More adversarial examples from the baseline model and CiiV counterpart.}
\label{fig:apx1}
\vspace{-3mm}
\end{wrapfigure}

In section~\ref{sec.4} of the original paper, we introduced a Confounded Toy (CToy) dataset to demonstrate the equivalence between the confounding effect and the adversarial perturbation. The proposed CToy is a three-way classification, containing triangles, squares, and circles. It has 10k/1k/1k images for train/val/test split, respectively. All samples are 64x64 colour images. Except for the causal geometries, there are also confounding patterns, \ie, red/blue/green blocks, with the size of 4x4 pixels. Different from the deterministic geometry, the color of each block is sampled from a biased distribution. For triangle, square, and circle images, each co-occurred block has 80\%/10\%/10\%, 10\%/80\%/10\%, and 10\%/10\%/80\% probability to be blue/green/red, respectively. Therefore, if the confounding distribution stays the same in both training and testing phases, these patterns are indeed ``predictive'' features. Yet, learning these confounding patterns would significantly reduce the generalization ability of the model, because there will always be samples that are dominated by rare color blocks, and they are also more brittle than geometry structures. The causal graph of the data generation procedure and more examples of CToy dataset are illustrated in Figure~\ref{fig:apx1} (a,b).

Based on the specifically designed CToy that only contains two patterns, causal shapes and confounding colors, we are able to understand which pattern causes the adversarial vulnerability. As we can see from Figure~\ref{fig:apx1} (c), adversarial examples of an $L_\infty$ PGD attack ($\epsilon$ is set to 128/255 for 100\% attacking success rate, so we can understand which pattern can successfully fool the model) that generated from a baseline DNN model were mainly erasing the original color blocks, \ie, the adversarial perturbation is indeed trying to maximize the tampered confounding effect. Specifically, the attacker changed the blue and red blocks in triangle and circle images to the green points. It even painted the entire square images to red. The confounding patterns were obviously tampered in these images while the causal geometries barely changed. On the other hand, the adversarial examples of the proposed CiiV model didn't change the overall colors too much, they directly modified the shapes. It proves that CiiV successfully prevents the model from learning confounding effects, and thus attacker can only poison the causal geometries. 

With the help of the CToy dataset, we are not only able to verify the proposed confounding theories for adversarial examples but also visualize the working mechanisms of the proposed CiiV framework, \ie, forcing the model to learn from causal patterns rather than the confounding colors.

\section{Details of The Proposed Causal Graph}
\label{sec:B}

\begin{figure}[t]
   \begin{minipage}[b]{1.0\linewidth}
   \centerline{\includegraphics[width=140mm]{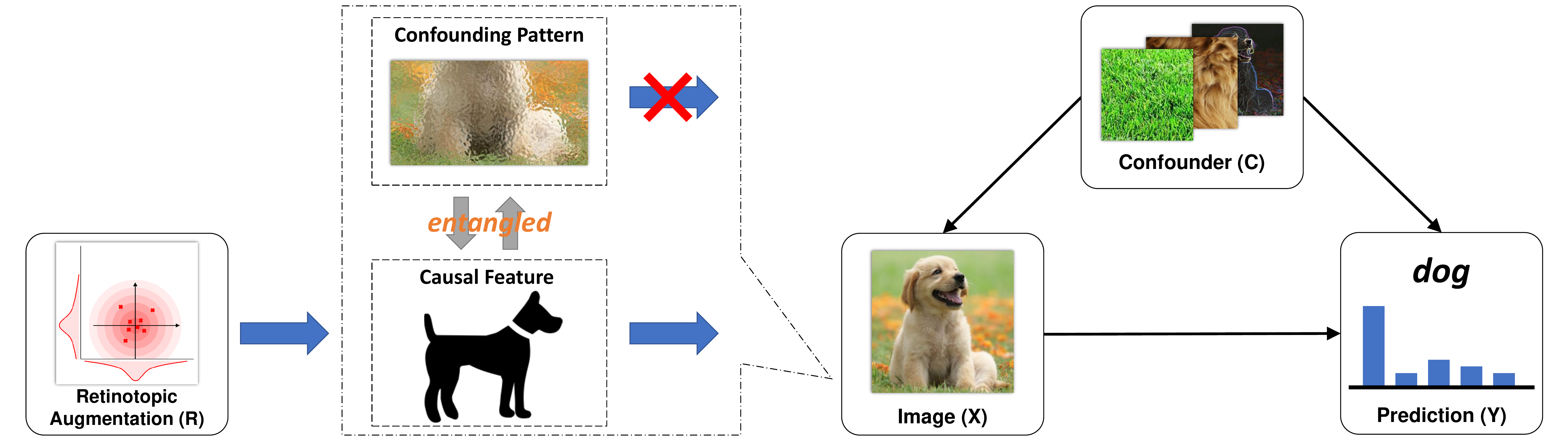}}
   \end{minipage}
   \caption{The details of the proposed causal graph for CiiV regularization and how confounding patterns cause the adversarial vulnerability.}
   \label{fig:apx2}
   \vspace{-5mm}
\end{figure}

In this paper, we firstly attribute the cause of non-robust features, which were originally introduced by \cite{ilyas2019adversarial} as an explanation of adversarial examples, to the ubiquitous confounding effect. But how do confounding patterns affect the learning of causal features and thus hurt the adversarial robustness? We believe the answer is the failure of feature disentanglement~\cite{yang2021causalvae, higgins2016beta}. As shown in Figure~\ref{fig:apx2}, a real-world image is usually composed of both concepts and contexts. Since those contexts often show statistical correlations with the causal concept, it's difficult to disentangle the concept from the context through pure observational data, \eg, the grass feature is usually co-occurred with the dog concept, but it's also shared by other outdoor images and absent in indoor dog images, so it's not a valid causal feature. Due to the unsuccessful feature disentanglement, adversarial perturbations that simply modify the grass texture would also lead to the collapse of dog feature, which eventually fool the predictor.

However, the feature disentanglement~\cite{yang2021causalvae, higgins2016beta} \textit{per se} is still an open question in machine learning. Otherwise, we only need to simply disentangle the robust and non-robust features then learn a classifier based on robust features. To tackle the adversarial vulnerability in practice, we need to bypass the trap of confounder disentanglement and seek help from the causal intervention without confounder observation, \ie, the instrumental variable estimation. As we introduced in section~\ref{sec.6}, there are two requirements for the choice of instrumental variable. The independence of $R$ can be directly guaranteed by the manual design of retinotopic sampling masks. To satisfy the second requirement that the effect of instrumental variable $R$ on $Y$ can only pass through the causal link $X\to Y$, we assume that causal features are global structures that change consistently across different retinotopic masks while the adversarial patterns are local impulses~\cite{du2020rain} that simply collapse after applying different retinotopic sampling. Note that this assumption limited the scope of our $C$ to those fragile confounding patterns, which is not trying to disentangle the semantically meaningful confounders.  Fortunately, those semantically meaningful confounders brought by the unbalanced dataset also won't be utilized as adversarial perturbations, \eg, the keyboard is usually co-occurred with the monitor and becomes a confounder of the latter, but the adversarial attack is obviously not allowed to create or erase a keyboard for the monitor image based on its definition. Therefore, our assumption still guarantees the proposed retinotopic sampling to be a valid instrumental variable in the adversarial robustness task.

\section{Details of The Retinotopic Augmentation}
\label{sec:C}

In this section, we will introduce the detailed implementation of retinotopic augmentation and the selection of its hyper-parameters. The proposed retinotopic augmentation layer $x_r=f(x,r)$ applies a centralized mask $r$ onto the image $x$, which imitates the biological retina that the central fovea has significantly denser photoreceptors than the peripheral. The mask $r$ is generated by a non-uniform spatial sampling, whose sampling probability decreases from a given center to the peripheries. We conjecture that human vision benefits from the visual attention and continuous eye movement~\cite{land2019eye} to implicitly apply diverse $r$ as the instrumental variable estimation. Intuitively, such a conjecture also explains why human can increase the recognition accuracy by continually gazing at different positions of an object, and why attention or focusing is so important in recognition.

\begin{wrapfigure}{r}{7.5cm}
\vspace{-3mm}
\includegraphics[width=7.5cm]{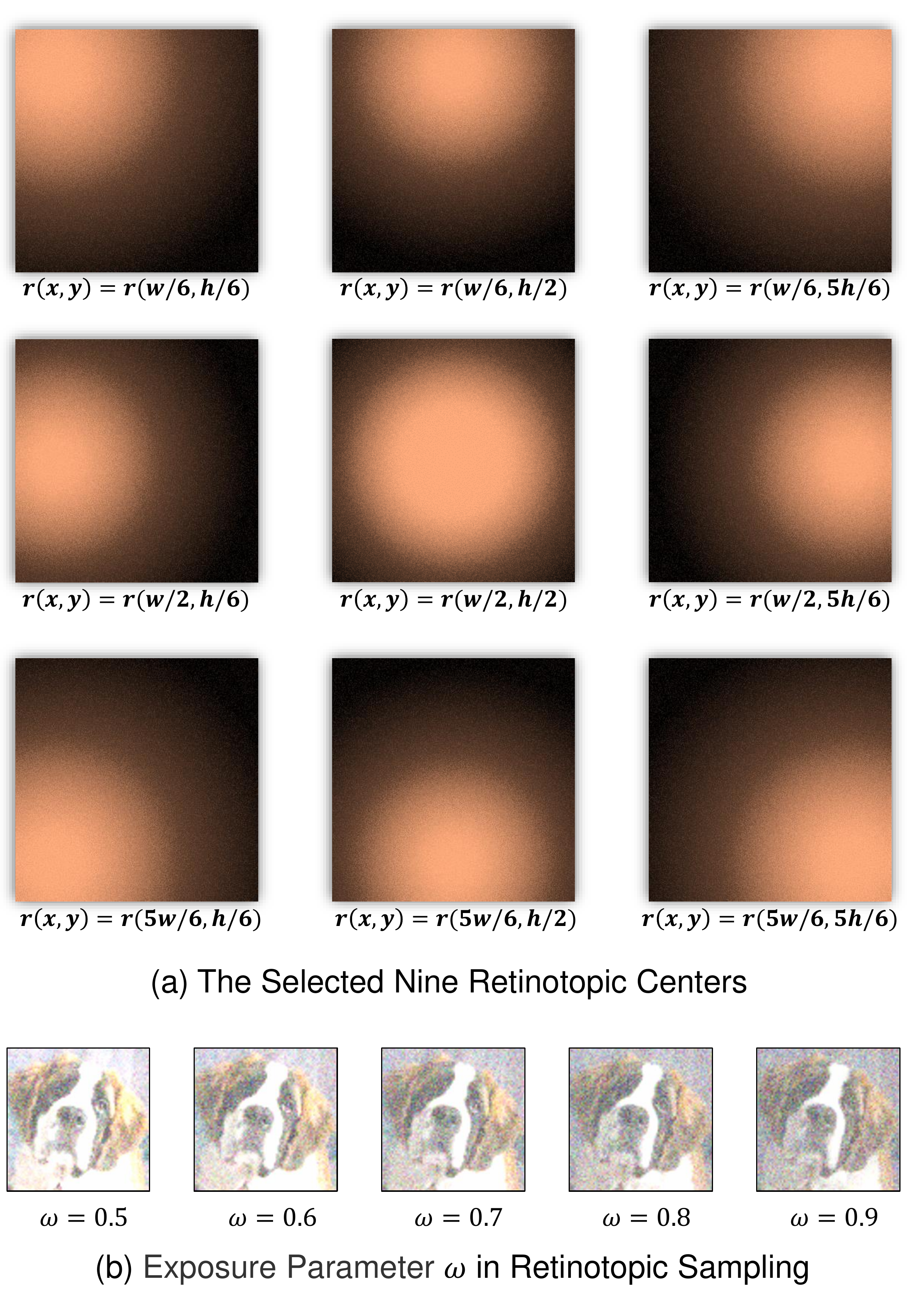}
\caption{(a) The selected 9 retinotopic centers used to generate $r$ in the proposed CiiV. (b) The effect of applying different exposure parameter $\omega$ before multiplying with the retinotopic sampling mask $r$}
\label{fig:apx3}
\vspace{-3mm}
\end{wrapfigure}

To conduct instrumental variable estimation, we adopt 9 sampling centers $(x, y)$ to generate different $r$. As illustrated in Figure~\ref{fig:apx3} (a), they are $(w/6, h/6)$,  $(w/6, h/2)$, $(w/6, 5h/6)$, $(w/2, h/6)$, $(w/2, h/2)$, $(w/2, 5h/6)$, $(5w/6, h/6)$, $(5w/6, h/2)$ and $(5w/6, 5h/6)$, where $w$ and $h$ are the width and height of each corresponding image. Note that the fixed retinotopic centers are only used to ensure the diversity of selected candidates, simply choosing 9 random centers could obtain very similar performances as shown in Table~\ref{tab:4}. Given the retinotopic center $(x,y)$, we define the retinotopic sampling mask $r$ as follows:
\begin{equation}
    r_{ij}(x,y) = g(\| (i,j)-(x,y)\|_2) + \varepsilon > \tau,
    \label{eq:sup1}
\end{equation}
where $i \in [0, w],j \in [0, h]$ are the indexes of image pixels, $g(\cdot)$ is a non-linear mapping that can be implemented by various functions, $\varepsilon$ is uniformly sampled from $[0,1]$, $\tau=0.9$ is the sampling threshold.  The spatial coverage $\alpha_r$ used in CiiV is defined as the coverage of the retinotopic mask $\sum r_{ij} / (w*h)$.

Note that the non-linear smoothing function $g(\cdot)$ of $r(x,y)$ can take various implementations, which won't affect the performances of the proposed CiiV too much as long as the sampling frequency decreases from the center $(x,y)$ to the peripheries as shown in Figure~\ref{fig:apx3} (a). We intuitively adopt a normalized mapping $g(z)=h((max(z)-z+\alpha)^{\gamma})$, $\alpha=10.0, \gamma=0.3, h(z)=z/max(z)$ as our default setting, and we further tested two simpler non-linear functions Candidate1: $g_1(z)=1.0-z/100$ and Candidate2: $g_2(z)=2.5/(0.5 \times z^{0.5})$. According to the experimental results in Table~\ref{tab:apx1}, different $g(\cdot)$ candidates perform very similarly under all attack settings, which proves that CiiV is not sensitive to the detailed implementations of $r(x,y)$. The main reason for us to choose a more complex implementation of $g(\cdot)$ is that it can dynamically fit the image size. The other two simpler functions $g_1(\cdot)$ and $g_2(\cdot)$ have to change parameters for different sizes of images, which is less convenient than our default $g(\cdot)$.

Since the proposed retinotopic augmentation aims to imitate the continuous observations in the human vision. The reaction of biological visual system to different light intensities is also important, which controls the amount of light absorbed by the retina. Therefore, given the retinotopic mask $r$ generated by $g(\cdot)$, the overall retinotopic augmented image $x_r$ can thus be constructed by $x_r=f(x,r) = 1/N \sum_i (r \odot ReLU(x + \varepsilon_i))$, where $\varepsilon$ is the parameter of exposure intensity uniformly sampled from $(-\omega, \omega)$ by $N$ times ($N$ and $\omega$ is set to 3 and 0.9, respectively, in our experiments),  $\odot$ denotes element-wise multiplication after normalizing the light intensity. The reason we introduce the function $ReLU(x + \varepsilon_i)$ is to find the best exposure ratio for a dataset. As we can see from Figure~\ref{fig:apx3} (b), the selection of exposure parameter $\omega$ can change the intensity of an observed image. The dark environment requires a smaller $\omega$, so we make it as a hyper-parameter for each dataset. We set $\omega$ to 0.9, 0.9, 0.8 for CIFAR-10, CIFAR-100, and mini-ImageNet, respectively. Intuitively, such a function imitates how human eyes react to different light intensities of the environment by controlling the amount of absorbed light. After multiplying with the retinotopic mask $r$, the proposed $x_r=f(x,r)$ simulates the signals perceived by the biological retina under different environments and focusing points, which continuously ``intervene'' the images observed by humans.

We also investigated hyper-parameters of retinotopic augmentation. As we can see from Table~\ref{tab:apx1}, there are trade-offs among different selections. Larger $\omega$ can capture more dark details at the cost of light details, and vice versa. To obtain the balanced results between clean images and adversarial examples, we chose the $\omega=0.9$ as our default setting in CIFAR-10. As to the parameter $N$ that is used to smooth the image after retinotopic augmentation, the larger $N$ we use the less distortion will be in the generated $x_r$. Therefore, larger $N$ can significantly increase the performance of clean images while smaller $N$ can increase the performance of adversarial examples by suppressing more confounding patterns. Although $N=1$ would obtain the best overall result, considering the fact that clean images occur more often than adversarial examples in real-world applications, we adopted $N=3$ as our default setting in all datasets.

\section{More detailed studies and experiments}
\label{sec:D}

In this section, we demonstrate additional studies and experiments on 1) several gradient-free attacks, and 2) more backbones.

\begin{wraptable}{r}{7.5cm}
\centering
\scalebox{0.7}
{
\begin{tabular}{c |c |c |c |c |c |c}
\hline
\hline 
Datasets & \multicolumn{6}{c}{CIFAR-10} \\
\hline
Attackers & Clean & FGSM & PGD-10 & AA-$l_{\infty}$ & AA-$l_2$  & Overall\\ 
\hline 
Default & 86.89 & 64.44 & 50.75 & 43.23 & 82.48 & 65.56 \\
Candidate1 & 87.03 & 64.53 & 50.37 & 42.46 & 82.86 & 65.45 \\
Candidate2 & 87.53 & 63.23 & 48.88 & 41.05 & 82.64 & 64.67 \\
$\omega=1.2$ & 86.50 & 64.96 & 52.04 & 46.33 & 82.18 & 66.40 \\
$\omega=1$  & 85.96 & 64.66 & 50.81 & 43.88 & 81.72 & 65.41 \\
$\omega=0.8$ & 87.21 & 64.19 & 49.97 & 41.72 & 82.74 & 65.17 \\
$\omega=0.6$ & 86.47 & 62.65 & 47.70 & 39.53 & 81.36 & 63.54 \\
$N=1$ & 82.44 & 69.40 & 59.17 & 57.25 & 78.16 & 69.28 \\
$N=2$ & 85.84 & 66.52 & 55.01 & 48.36 & 81.36 & 64.42 \\
$N=4$ & 87.71 & 63.77 & 48.39 & 40.14 & 83.06 & 64.61 \\
$N=5$ & 88.29 & 62.22 & 46.35 & 37.87 & 83.92 & 63.73 \\
\hline
\hline
\end{tabular}
}
\caption{The performances of CiiV on CIFAR-10 using different designs of function $g(\cdot)$ to generate retinotopic sampling mask $r$, and different hyper-parameters $\omega$ and $N$ to generate $x_r$.}
\label{tab:apx1}
\end{wraptable}

\begin{table*}
\centering
\scalebox{0.7}
{
\begin{tabular}{c ||c |c |c |c |c ||c |c |c |c |c}
\hline
\hline
Datasets & \multicolumn{5}{c||}{CIFAR-10} & \multicolumn{5}{c}{CIFAR-100} \\ 
\hline 
Attackers & Clean & GN & UN & SPSA & BFS & Clean & GN & UN & SPSA & BFS\\ 
\hline 
Baseline & 94.42 & 72.05 & 74.66 & 68.60 & 35.77 & 74.53 & 31.44 & 34.46 & 27.57 & 10.39 \\
mixup & 95.31 & 76.02 & 78.77 & 71.87 & 39.82 & 77.32 & 40.21 & 43.69 & 36.27 & 18.35 \\
BPFC  & 90.21 & 88.90 & 89.02 & 88.91 & 79.48 & 61.48 & 60.47 & 60.53 & 60.30 & 52.11 \\
RS & 83.44 & 83.22 & 83.16 & 83.35 & 82.97 & 54.63 & 54.46 & 54.28 & 54.53 & 54.41 \\
(ours) CiiV & 86.89 & 86.47 & 86.61 & 86.75 & 85.60 & 58.88 & 58.19 & 58.75 & 58.63 & 57.30 \\
(ours) CiiV+mixup & 87.14 & 86.71 & 87.11 & 87.07 & 86.23 & 56.90 & 56.41 & 56.65 & 56.70 & 55.91 \\
(ours) CiiV+RandAug & 89.12 & 88.36 & 88.59 & 89.00 & 87.64 & 59.26 & 58.82 & 58.71 & 59.21 & 57.85 \\
\hline
AT$_{FGSM}$ & 84.52 & 83.02 & 83.32 & 82.98 & 77.86 & 51.99 & 51.05 & 51.27 & 51.13 & 45.08 \\
AT$_{PGD-10}$ & 83.94 & 82.49 & 82.70 & 82.59 & 77.33 & 56.48 & 54.64 & 54.97 & 54.82 & 47.83 \\
(ours) CiiV+AT$_{FGSM}$ & 83.67 & 83.10 & 83.20 & 83.37 & 82.34 & 53.83 & 53.35 & 53.77 & 53.69 & 52.53 \\
(ours) CiiV+AT$_{PGD-10}$ & 81.35 & 80.74 & 81.02 & 81.09 & 80.22 & 51.73 & 51.16 & 51.43 & 51.39 & 50.12 \\
\hline
\hline
\end{tabular}
}
\caption{Gradient-free attacks on CIFAR-10 and CIFAR-100. The upper half contains the AT-free defenders while the bottom half reports the AT-involved defenders.}
\label{tab:apx2}
\end{table*}

\begin{table*}
\centering
\scalebox{0.63}
{
\begin{tabular}{c ||c |c |c |c |c |c ||c |c |c |c |c |c}
\hline
\hline
Datasets & \multicolumn{6}{c||}{CIFAR-10} & \multicolumn{6}{c}{CIFAR-100} \\ 
\hline 
Attackers & Clean & FGSM & PGD-10 & AA-$l_{\infty}$ & AA-$l_2$  & Overall & Clean & FGSM & PGD-10 & AA-$l_{\infty}$ & AA-$l_2$ & Overall\\ 
\hline 
(VGG13) Baseline & 90.20 & 10.48 & 0.0 & 0.0 & 0.21 & 20.18 & 66.05 & 3.14 & 0.0 & 0.0 & 0.05 & 13.85 \\
(VGG13) CiiV & 83.44 & 58.75 & 43.62 & 36.98 & 79.08 & 60.37 & 50.62 & 32.83 & 25.58 & 24.91 & 47.37 & 36.26 \\
(VGG13) CiiV+RandAug & 83.91 & 60.60 & 45.54 & 40.51 & 80.63 & 62.04 & 52.72 & 33.24 & 26.66 & 25.76 & 48.64 & 37.40 \\
\hline
(WRN34-10) Baseline & 94.93 & 32.09 & 0.02 & 0.0 & 0.0 & 25.41 & 77.74 & 9.73 & 0.15 & 0.0 & 0.0 & 17.52 \\
(WRN34-10) CiiV & 87.25 & 59.50 & 43.89 & 38.24 & 82.82 & 62.34 & 60.84 & 34.04 & 25.58 & 25.23 & 57.45 & 40.63 \\
(WRN34-10) CiiV+RandAug & 88.68 & 64.02 & 49.23 & 44.52 & 83.59 & 66.01 & 63.23 & 38.30 & 28.82 & 27.38 & 59.81 & 43.51 \\
\hline
\hline
\end{tabular}
}
\caption{The performances of Baseline, CiiV, and CiiV+RandAug using different backbones.}
\label{tab:apx3}
\end{table*}

According to \cite{carlini2019evaluating}, some flawed defenders may fail in gradient-free attacks. Therefore, we further investigated four gradient-free attackers: 1) GN (Gaussian Noise), 2) UN (Uniform Noise), 3) SPSA~\cite{uesato2018adversarial}, and 4) BFS (Brute-Force Search)~\cite{carlini2019evaluating}. Since gradient-free attacks are supposed to be much weaker than gradient-based attacks, we increased the budget $\varepsilon$ to 16/255 for all four gradient-free attackers under $l_\infty$ constraint. To be specific, GN and UN add gaussian and uniform noises, respectively, to input images. BFS ran 100 times of GN and reported the most vulnerable adversarial examples. As to the SPSA, it conducted numerical approximation of gradients to circumvent the potential gradient masking, the hyper-parameters were set as $\delta$=0.1, step=20, lr=0.1, batch size=16. According to the experiments in Table~\ref{tab:apx2}, all the gradient-free attackers were significantly weaker than the gradient-based attackers as we expected even with the doubled attacking budget, proving that the proposed CiiV won't be more vulnerable under gradient-free attacks. 

We also applied the proposed CiiV and its combination with Random Augmentation, \ie, the AT-free versions of defenders, into other backbones, \eg, VGG13~\cite{simonyan2014very} and WRN34-10~\cite{zagoruyko2016wide}. As we can see from Table~\ref{tab:apx3}, the proposed CiiV and its variants consistently increased the adversarial robustness under different backbone models.

\end{document}